\documentclass[pdflatex,sn-apa]{sn-jnl}

\usepackage{graphicx}%
\usepackage{multirow}%
\usepackage{amsmath,amssymb,amsfonts}%
\usepackage{amsthm}%
\usepackage{mathrsfs}%
\usepackage[title]{appendix}%
\usepackage{colortbl}%
\usepackage{xcolor}%
\usepackage{textcomp}%
\usepackage{manyfoot}%
\usepackage{booktabs}%
\usepackage{algorithm}%
\usepackage{algorithmicx}%
\usepackage{algpseudocode}%
\usepackage{listings}%
\usepackage{wasysym}%
\usepackage{textcomp}%
\usepackage{tabularx}%
\usepackage{enumitem}%
\usepackage{caption}%
\usepackage{subcaption}%
\usepackage{tablefootnote}%
\usepackage{float}%

\theoremstyle{thmstyleone}%
\theoremstyle{thmstyletwo}%

\theoremstyle{thmstylethree}%

\begin{document}

\title[]{Crosslingual Optimized Metric for Translation Assessment of Indian Languages}


\author*[1]{\fnm{Arafat} \sur{Ahsan}}\email{arafat.ahsan@iiit.ac.in}

\author[1]{\fnm{Vandan} \sur{Mujadia}}\email{vandan.mu@research.iiit.ac.in}

\author[1,2]{\fnm{Pruthwik} \sur{Mishra}}\email{pruthwikmishra@aid.svnit.ac.in}

\author[1]{\fnm{Yash} \sur{Bhaskar}}\email{yash.bhaskar@research.iiit.ac.in}

\author[1]{\fnm{Dipti Misra} \sur{Sharma}}\email{dipti@iiit.ac.in}

\affil[1]{\orgdiv{Language Technologies Research Centre}, \orgname{International Institute of Information Technology}, \orgaddress{\city{Hyderabad}, \country{India}}}

\affil[2]{\orgdiv{Department of AI}, \orgname{Sardar Vallabhbhai National Institute of Technology}, \orgaddress{\city{Surat}, \country{India}}}


\abstract{Automatic evaluation of translation remains a challenging task owing to the orthographic, morphological, syntactic and semantic richness and divergence observed across languages. String-based metrics such as BLEU have previously been extensively used for automatic evaluation tasks, but their limitations are now increasingly recognized. Although learned neural metrics have helped mitigate some of the limitations of string-based approaches, they remain constrained by a paucity of gold evaluation data in most languages beyond the usual high-resource pairs. In this present work we address some of these gaps. We create a large human evaluation ratings dataset for 13 Indian languages covering 21 translation directions and then train a neural translation evaluation metric named Cross-lingual Optimized Metric for Translation Assessment of Indian Languages (COMTAIL) on this dataset. The best performing metric variants show significant performance gains over previous state-of-the-art when adjudging translation pairs with at least one Indian language. Furthermore, we conduct a series of ablation studies to highlight the sensitivities of such a metric to changes in domain, translation quality, and language groupings. We release both the COMTAIL dataset and the accompanying metric models.}




\maketitle

\section{Introduction}
\label{sec:intro}

Translation is a chimera and its evaluation no less a daunting task. Viewed retrospectively, both tasks appear to undergo paradigmatic shifts in close succession. The early rule-based machine translation (MT) systems were solely reliant on human assessment methods that assessed MT outputs against an adjectival scale for Adequacy, Fluency, and Comprehensibility  \citep{alpac1966, white1994arpa}. The turn to data-intensive statistical systems for machine translation \citep{brown1993mathematics} and their subsequent popularity \citep{koehn2003statistical} was soon followed by the appearance of automatic evaluation metrics, such as the n-gram overlap-based metric BLEU \citep{papineni2002bleu}. BLEU undoubtedly brought scalability and standardization to the evaluation task, but it was not to be the panacea being sought \citep{callison2006re}. Translation Edit Rate \citep{snover-etal-2006-study} and later character n-gram {F}-score \citep{popovic-2015-chrf} were proposed to address some of the gaps that BLEU left in its wake. 

The story then turned neural, and with it the MT task experienced a never before leap in quality that addressed long-standing issues around reordering, availability of longer context windows, and handling of unseen words and phrases  \citep{bahdanau2015neural, vaswani2017attention}. Evaluation metrics were not far behind. Contextual embedding-based models were soon utilized \textit{with} \citep{sellam-etal-2020-bleurt, rei-etal-2020-comet} or \textit{without} \citep{zhang-etal-2020-bertscore} human ratings data to produce neural metrics that appeared to fare much better than surface metrics when correlated with human evaluations \citep{mathur-etal-2020-results}. 

In parallel, human evaluation methods were also evolving along with the changes in automated metrics. The Workshop on Machine Translation (WMT) Shared Tasks, for a period of almost two decades since their inception, mirror this shift. This entire period can broadly be categorized as three distinct phases: the first phase, where traditional adjectival rating-based evaluation methods along with relative rankings made up the majority of human gold judgements \citep{bojarTenYearsWMT2016}; the second phase, starting 2016 was marked by the adoption of Direct Assessment (DA) methodology employing large scale crowd-sourced human ratings, later tethered to a Scalar Quality Metric (SQM) scale \citep{grahamContinuousMeasurementScales2013, graham2017da, bojar2016findings, bojar2017findings}; and we may now be at the start of a third phase, with the preferred methodologies reliant on close human error analysis of translation outputs \citep{freitag2021wmt21metrics, kocmi2024esa}, employing much expanded and somewhat standardized error typologies like MQM \citep{lommel2014multidimensional}.

Unlike earlier evaluation metrics such as BLEU, TER, or chrF, which were string-overlap measures requiring reference translations but not large amounts of human judgment data, much of the current work is data-hungry. To our knowledge, the largest dataset available for metric training is in the form of concatenated crowd-sourced DA ratings from the WMT shared tasks that amount to approximately 1.29 million items\footnote{\url{https://huggingface.co/datasets/RicardoRei/wmt-da-human-evaluation}}. However, the language coverage is limited, with only 41 unique language directions covered, with most language pairs being English centric. Moreover, the quality of these crowd-sourced ratings has also been questioned \citep{freitag2021experts, graham2017can}. 

However, a further advantage of neural metrics is the possibility of reference-less evaluation \citep{specia2010machine}. This is important in light of the fact that reference \textit{gold} translations are often of poor quality, especially for translations into low-resource languages \citep{toral2018attaining, freitag2021experts, abdulmumin-etal-2024-correcting}. Any method that eschews this reliance on gold reference translations stands a good chance of not only improved correlations with human judgements, but also in its applicability to production systems where  continuous monitoring of system quality may be required. These are known as Quality Estimation (QE) metrics and have been a regular feature of WMT Shared Tasks \citep{mathur2020results}. More recently, we have also seen the advent of Large Language Model (LLM) based evaluation methods such as GEMBA, GEMBA-MQM and TOWER \citep{freitag2022gemba, kocmiGEMBAMQMDetectingTranslation2023, alvesTowerOpenMultilingual2024}.

It is with this background that we place our current work into this most recent evolutionary wave of translation evaluation metrics and methodologies. We present \textbf{COMTAIL}: \textbf{C}ross-lingual \textbf{O}ptimized \textbf{M}etric for \textbf{T}ranslation \textbf{A}ssessment of \textbf{I}ndian \textbf{L}anguages), a neural translation evaluation metric trained on a significant volume of Indian language translation ratings data, that we create. After averaging over 1 million individually rated items and filtering them for quality, the ratings dataset that we release contains $221941$ items across all its splits. These ratings have been collected in 21 translation directions covering 13 Indian languages on a diverse set of translation outputs. The metric models trained on this data show improved performance when adjudging translation pairs that contain an Indian language. Furthermore, we also present results from an LLM-based evaluation task, with models we fine-tune. We also conduct a series of ablation studies to uncover the sensitivities of these models to changes in volume, domain, quality, and language groupings. We release both the data and the metric models\footnote{\url{https://github.com/aenaliph/COMTAIL}}.

In this work, we address the following research questions:

\begin{enumerate}[label=\textbf{RQ\arabic*}, leftmargin=2.5em]
    \item How effective are current neural machine translation metrics in evaluating Indian language translation directions?
    \item Can large language models (LLMs) serve as reliable evaluators of translation quality for Indian languages?
    \item How do reference-based and reference-less evaluation compare in the low-resource Indian language setting?
    \item How much data is required to reach stable performance when building automatic evaluation metrics?
    \item How do data domain shifts impact trained neural metrics?
    \item How does linguistic relatedness affect trained neural metrics?
    \item How do these metrics fare on a variety of linguistic phenomena?
\end{enumerate}

Additionally, we also present in detail our methodology in creating this ratings dataset and highlight our learnings from experiment design to pitfalls inherent when undertaking crowd-sourced tasks.

The rest of this paper is organized as follows. In Section \ref{sec:related} we discuss past efforts that have touched upon some of the same issues and questions that we raise in our current work. In Section \ref{sec:data} we discuss our methodology in creating the COMTAIL ratings dataset, from rater selection to task execution. We then describe our experimental setup in section \ref{sec:experiment}. This is followed by a discussion of results in Section \ref{sec:results} and we conclude with Section \ref{sec:conclusion}.

\section{Related Work}
\label{sec:related}

\subsubsection*{Human Evaluation Data}
A large portion of the human evaluation ratings available today originate from the WMT Translation Shared Tasks, the conventional venue to operationalize, refine, and report new methodologies. Particularly noteworthy are its findings from 2017 that introduce Direct Assessment evaluations on a 0--100 scale and detail the quality control methods followed to filter crowd-sourced ratings \citep{bojar2016findings}. In 2022 WMT combined the continuous Direct Assessment method with 0--6 point Scalar Quality Metric, which was said to help stabilize scores across annotators in comparison to DA \citep{kocmiFindings2022Conference2022}. \cite{freitag2021experts} argue instead for the use of MQM or SQM based evaluations conducted by professional evaluators in view of low correlations on WMT data rated by crowd workers.

\subsubsection*{Indian Languages}
However, Indian language representation remains quite low in these datasets with only 4 Indian languages listed in the combined DA dataset mentioned previously. \cite{saibIndicMTEvalDataset2023} somewhat address the Indian language representation issue by creating an MQM dataset consisting of 7000 fine-grained annotations for 5 Indian languages. \cite{singhHowGoodZeroShot2024} extend this to another 4 low-resource Indian languages by adding another 1000 MQM annotations. Both of these efforts primarily focus on MQM annotations,  although a direct score on a scale of 0--25 is also elicited along with the annotations. In contrast, our work is based on the DA+SQM methodology covering 13 Indian languages in 21 translation directions. 

\subsubsection*{Neural Metrics}
\cite{rei-etal-2020-comet} introduced the COMET framework that utilized cross-lingual pre-trained models and exploited information from both the source as well as the target reference translation. The resulting evaluation models saw a significant improvement in human correlation values. A number of iterations on this initial architecture produced bigger and better models \citep{reiCOMET22UnbabelIST20222022, kocmiMSCOMETMoreBetter2022}. Reference-less QE models built with a similarly adapted architecture also proved to correlate highly with human judgements, becoming competitive with and an alternative to their reference-based counterparts \citep{rei2021references, rei2022cometkiwi}. Some of these underlying COMET variants, trained with DA and MQM annotations, were used to fine-tune with small amounts of Indian language evaluation data and have shown improved correlations for those languages \citep{saibIndicMTEvalDataset2023}. 

Our effort and experiments with the COMET architecture and models extends past work in the following ways: we use a large Indian language evaluation dataset, the largest so far; we explore both training from scratch and fine-tuning strategies; we perform these for both reference-based and reference-less COMET architectures; and we report an extensive series of ablation experiments that uncover the effects of data volume, domain, quality and language groupings on these models.

\subsubsection*{LLM-based Metrics}
Large Language Models are also becoming increasingly popular for translation related tasks owing to the fact that the underlying model can be adapted to a number of different tasks found in a typical translation workflow \citep{alvesTowerOpenMultilingual2024}. Models such as GEMBA-DA and GEMBA-MQM claim state of the art performance on some evaluation datasets \citep{kocmiLargeLanguageModels2023, kocmiGEMBAMQMDetectingTranslation2023}. However, these particular past efforts relied on closed proprietary models. \cite{mujadiaLargeLanguageModel2024} fine-tune a \textit{LLaMA2-13B} model on Indian language Quality Estimation data from WMT and report correlation values at par with state of the art COMET-based metrics. We report results obtained by fine-tuning a \textit{Llama3.1-8B} model on our dataset for both reference-based and reference-less evaluation tasks.

\section{Dataset Creation}
\label{sec:data}

\subsection{Languages}

In India, a large scale national effort is currently underway towards addressing language technology and translation needs for all of the 22 constitutionally recognized Indian languages\footnote{\url{https://bhashini.gov.in}}. Our work is a contribution towards this larger effort, but here we narrow our focus to 13 Indian  languages, owing to resource and budgetary constraints. 

Table \ref{tbl:dataset-languages} shows the language pairs that make up our dataset. Pairing 2 source languages, English and Hindi with 13 targets yields a total of 22 translation directions. Also listed are the standard \textit{ISO 639-3} language codes, their respective language families and scripts. While we do not venture to classify each language into high or low resource, it is noteworthy that Kashmiri, Dogri, and Sindhi currently have little to almost no digital presence.

\begin{table}[h]
\begin{minipage}\textwidth
\caption{Language coverage of our dataset. Translations from two source languages, English and Hindi paired with 13 unique Indian target languages, making a total of 22 unique translation directions. Also listed are the ISO~639-3 codes, language family, and the predominant script in use for each language. '–' implies that we did not create data for that particular source-target combination.}
\label{tbl:dataset-languages}
\centering
\begin{tabular*}{\textwidth}{@{\extracolsep{\fill}}l l l l l}
\toprule
\multicolumn{2}{c}{\textbf{Source Languages}} & \textbf{Code} & \textbf{Family} & \textbf{Script} \\
\cmidrule(lr){1-2}
\textbf{English $\rightarrow$} \{...\} & \textbf{Hindi $\rightarrow$} \{...\} & & & \\
\midrule
Bengali   & Bengali   & ben & Indo-Aryan & Brahmic \\
Gujarati  & Gujarati  & guj & Indo-Aryan & Brahmic \\
Hindi     & –         & hin & Indo-Aryan & Brahmic \\
Kannada   & Kannada   & kan & Dravidian  & Brahmic \\
Kashmiri  & –         & kas & Indo-Aryan & Perso-Arabic \\
Marathi   & Marathi   & mar & Indo-Aryan & Brahmic \\
Odia      & Odia      & ori & Indo-Aryan & Brahmic \\
Punjabi   & Punjabi   & pan & Indo-Aryan & Brahmic \\
–         & Sindhi    & snd & Indo-Aryan & Perso-Arabic \\
Tamil     & Tamil     & tam & Dravidian  & Brahmic \\
Telugu    & Telugu    & tel & Dravidian  & Brahmic \\
Urdu      & Urdu      & urd & Indo-Aryan & Perso-Arabic \\
–         & Dogri     & doi & Indo-Aryan & Brahmic \\
\bottomrule
\end{tabular*}
\end{minipage}
\end{table}

\subsection{Raters}
\label{sec:data-raters}
One of the key aspects to consider when undertaking a human translation evaluation task is the quality of evaluators. There are two primary possibilities when assembling teams for such a task: evaluation at scale via online crowd-sourcing tasks with Quality Control (QC); or evaluation by a select group of professional translators and evaluators. It has previously been shown that professional evaluators tend to produce less noisy ratings \citep{freitag2021experts}, but such a setting is harder to scale. For our study we chart a middle path by incorporating ideas from both approaches. Our eventual raters turn out to be a mix of professionals and freelancers, carefully curated via selection criteria we describe below.

Human Evaluation of machine translation (MT) outputs can consist of the following: a \textit{source} that has been translated; a \textit{reference}, that may or may not be available, and which is usually human translated and assumed to be of good quality; and finally, a \textit{hypothesis} that is to be analysed or rated. This analysis or rating is often performed in light of the \textit{source} itself (bilingual evaluation) or solely against the \textit{reference} (monolingual evaluation); as to which is chosen depends on the availability and competence of bilingual or monolingual evaluators. 

Table \ref{tbl:dataset-eval-type} lists the evaluation possibilities. The main advantage of monolingual evaluation is access to a large pool of target language evaluators, in contrast to the scarcity of bilingual evaluators, particularly for less common source–target language pairs. However, the former's advantage hinges on a crucial aspect, that of good quality human translated \textit{references} being available against which the \textit{hypotheses} are to be compared. On the other hand, bilingual evaluation is primarily dependent on the competency of the evaluator in both the source and target languages. While native proficiency and competence can be assumed when the target language is the native language of the evaluator, the same cannot be assumed when it comes to source language competence. With this in mind we chose to administer a \textit{Reading Comprehension} test to all evaluators who expressed an interest in the task by registering with us.

\begin{table}[ht]
\caption{Evaluation Types}
\label{tbl:dataset-eval-type}
\centering
\begin{tabular*}{\textwidth}{@{\extracolsep{\fill}}l l l}
\toprule
Evaluator &  Evaluation Pair & Expected Proficiency \\
\midrule
Monolingual &        \textit{hypothesis}, \textit{reference} & Target Language\\
Bilingual &        \textit{source}, \textit{hypothesis} & Source \textit{AND} Target Language\\
\bottomrule
\end{tabular*}
\end{table}

\subsubsection*{Reading Comprehension Test}
We prepared a reading comprehension test with 10 questions each for both of the source languages. Prospective raters were asked to read short passages and answer questions related to them. The English test consisted of four passages and ten questions. The first three passages were in the general domain and were taken from \textit{RACE-C}, a multiple choice reading comprehension dataset collected from college-level English examinations in China \citep{liang2019new}, while the last passage was from \textit{CliCR} \citep{vsuster2018clicr}, a clinical dataset, to represent a health domain subset that is part of our dataset. The questions were a mix of \textit{Cloze} and \textit{Multiple Choice} formats, all requiring the evaluator to select a correct option out of the given choices.

The Hindi Reading Comprehension test comprised of five passages and ten questions. One passage was taken from a public website that hosts preparatory material related to Central Board of Secondary Examination's (CBSE) Class 12 exams in India, equivalent to the 12th Grade in United States and A-levels in the United Kingdom. The remaining passages were taken from Hindi news stories published online and the questions were formulated by one of the Authors, a native speaker of the language. These again were a mix of \textit{Cloze} and \textit{Multiple Choice} formats, requiring the selection of one option out of many. All of the passages in this case were from the general domain.

A sample passage for each source language and associated questions for the passage are reproduced in Appendix \ref{app:details-rc-sample}.

Figure \ref{fig:rc-distribution} shows the performance of raters on the Reading Comprehension Test. Score distributions for each of the source languages and the corresponding target languages are shown. Tables \ref{tbl:dataset-eng-rc-scores} and \ref{tbl:dataset-hin-rc-scores} show the number of raters per language pair who took the test and those who passed it at a score threshold of $6$/$10$. These were the final numbers that participated in the actual rating task. 

\begin{figure}
  \centering
  \includegraphics[width=0.9\textwidth]{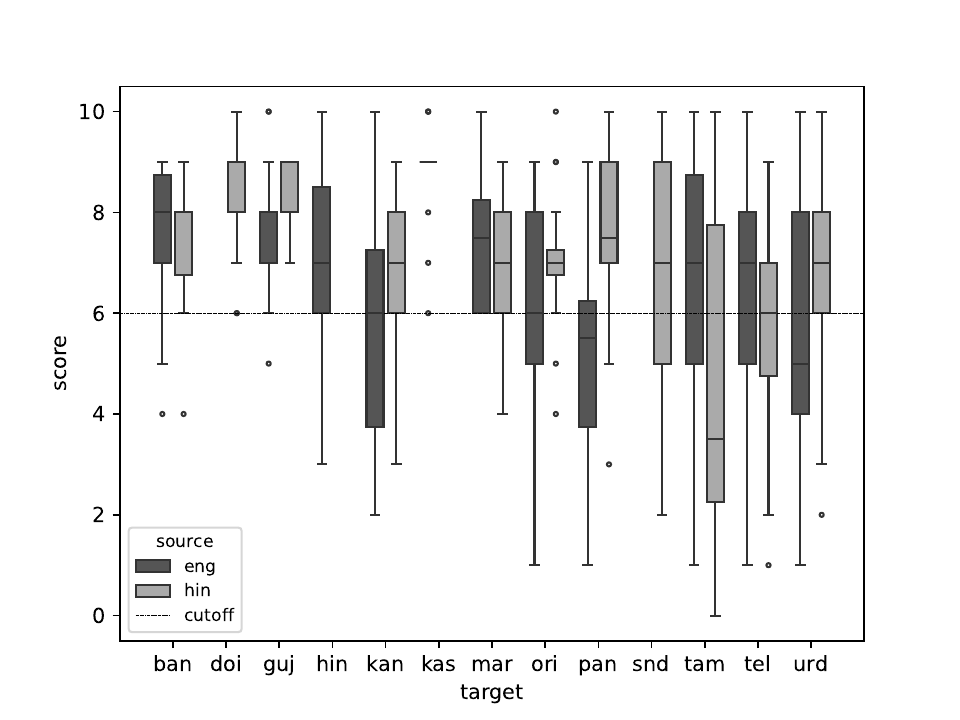}
  \caption{Distribution of Reading Comprehension Test scores for each target for the two source languages. The \textit{dash-dot} line shows the cut-off threshold of $6$/$10$. All raters at and above the threshold were shortlisted for the rating task.}
  \label{fig:rc-distribution}
\end{figure}

\begin{table}[h]
\begin{minipage}{0.45\textwidth}
\caption{Raters who made the cut-off for English Reading Comprehension and means by target language}
\label{tbl:dataset-eng-rc-scores}
\centering
\begin{tabular*}{\textwidth}{@{\extracolsep{\fill}}l l l l} 
\toprule
source & target & passed & Mean \\
\midrule
eng & ban & 16/18 & 7.56 \\
eng & guj & 16/17 & 7.76 \\
eng & hin & 35/43 & 7.00 \\
eng & kan & 14/24 & 5.79 \\
eng & kas & 13/13 & 8.77 \\
eng & mar & 16/16 & 7.50 \\
eng & ori & 11/17 & 6.06 \\
eng & pan & 12/24 & 5.21 \\
eng & tam & 27/38 & 6.63 \\
eng & tel & 41/57 & 6.77 \\
eng & urd & 13/27 & 5.74 \\
\midrule
eng & all & 214/294 & 6.80 \\
\bottomrule
\end{tabular*}
\end{minipage}
\centering
\hfill
\begin{minipage}{0.45\textwidth}
\caption{Raters who made the cut-off for Hindi Reading Comprehension and means by target language}
\label{tbl:dataset-hin-rc-scores}
\begin{tabular*}{\textwidth}{@{\extracolsep{\fill}}l l l l}
\toprule
source & target & passed & Mean \\
\midrule
hin & ban & 7/8 & 7.25 \\
hin & doi & 14/14 & 8.14 \\
hin & guj & 17/17 & 8.24 \\
hin & kan & 13/16 & 6.75 \\
hin & mar & 22/24 & 7.21 \\
hin & ori & 18/20 & 7.05 \\
hin & pan & 24/28 & 7.61 \\
hin & snd & 30/41 & 7.00 \\
hin & tam & 2/6 & 4.67 \\
hin & tel & 19/28 & 5.82 \\
hin & urd & 17/21 & 6.71 \\
\midrule
hin & all & 183/223 & 6.95 \\
\bottomrule
\end{tabular*}
\end{minipage}
\end{table}

\subsection{Rating Scale}
As we briefly touched upon earlier, human  evaluation methodologies have co-evolved along with newer metrics and models. As per the latest WMT findings, Error Span Annotation (ESA) which involves highlighting/marking errors without classifying them into different error types is the current preferred human evaluation method at their shared tasks \citep{kocmiErrorSpanAnnotation2024}. However, along with ESA evaluators continue to assign a 0--100 score similar to DA \citep{kocmiFindingsWMT24General2024}. Prior to this in 2022 and 2023, it was DA in combination with SQM that was deployed for human evaluation \citep{kocmiFindings2022Conference2022, kocmiFindings2023Conference2023}. We chose the \textbf{DA+SQM} method as the ESA methodology is still quite new and unproven at this time. We use source-DA which requires bilingual evaluators and we tether it to a 7 point SQM scale.

\subsection{Data Preparation}
\label{sec:dataset-prep}
Data used for the ratings task were pooled from multiple sources for each language pair. We utilized available parallel corpora, which facilitated both the evaluation of existing human translation quality and the subsequent use of these translations for benchmarking and training reference-based metrics. The data sources were a mix of existing publicly available corpora, especially, for the English source \citep{ramesh2022samanantar}, and an as yet unreleased Indian Language to Indian Language translation dataset for the Hindi source. The domain information for the data were coded and mapped to three domains: \textit{general}, \textit{governance} and \textit{health}; their proportions varied across language pairs depending on the availability of domain data in each of the language pairs.

\subsubsection*{Source Sentences}
In addition to the domain label, the source sentences for English and Hindi were also categorized into four length buckets: 0–10, 10–20, 20–35, and 35–100 and the actual data used in the ratings task were proportionally sampled from this larger assembled pool.

For each translation direction, we had initially planned to pair 3000 unique source sentences with multiple target hypothesis and rate the ensuing items, but the final coverage turned out to be less than 3000 unique sentences per pair and was highly dependent on the pace at which the actual task progressed and the number of raters available for that pair.

\subsubsection*{Target Hypotheses}
Once we had this raw source data pool as described in the previous section, we then set about creating a variety of translation hypothesis that were to be rated. We chose outputs from a diverse mix of Machine Translation (MT) engines representing a diversity of data and architectures. The available MT engines varied across language pairs, especially for the extremely low-resourced languages in our study (Dogri, Kashmiri, Sindhi).

In addition to the MT engines, we also included the original \textit{gold} reference from our data as a hypothesis to be rated. We also introduced two types of perturbations on this reference. These gold and perturbed targets were meant to act both as Quality Control items (as we will see in a later section), as well as help in diversifying the distribution of ratings across quality levels.

We thus ended up with the following 10 translation hypotheses including MT engines, gold references, and perturbed outputs: \textit{GPT3.5, Google, IndicTrans2, MS\_Bing, Seamless, gold, oldX, perturb, perturbed\_multiple, versionvN}. We describe each in turn:

\begin{itemize}[label={},leftmargin=0pt,itemsep=1em]

\item
\textbf{GPT-3.5}: GPT-3.5 is a decoder only transformer model from the GPT-3 family of OpenAI models and was the representative LLM-based translator in our experiments \citep{brown2020language}. More specifically, we used the \textit{GPT-3.5-turbo-0125} variant.\footnote{\url{https://platform.openai.com/docs/models/gpt-3-5-turbo}} The translations using OpenAI's API were generated during the period of 17 February 2024 to 19 February 2024. The following prompt template was used to generate the translations for each source and target pair:

\begin{verbatim}

    Translate the following {src_lang} text to {tgt_lang}
    {src_lang}: {src_sent}
\end{verbatim}

\item 
\textbf{Google Translate}: Google's Translate\footnote{\url{https://translate.google.co.in}} is a publicly available online translation engine. The architecture details and model size are not in the public domain. The translations for each source target pair were generated between 17 February 2024 to 19 February 2024 using the Google Translate API.
    
\item 
\textbf{IndicTrans2}:
IndicTrans2 Models are transformer-based multilingual models trained on 22 Indian Languages \citep{gala2023indictrans}. We utilize the en-indic\footnote{\url{https://huggingface.co/ai4bharat/indictrans2-en-indic-1B}} and indic-indic\footnote{\url{https://huggingface.co/ai4bharat/indictrans2-indic-indic-1B}} \textit{1B} models available via Huggingface. 

\item 
\textbf{versionvN}:
VersionvN refers to a multilingual transformer based in-house machine translation model that has been trained on data for 22 scheduled Indian languages \footnote{\url{https://ssmt.iiit.ac.in/onemt}}. 

\item 
\textbf{MS\_Bing}:
Microsoft's Bing Translator\footnote{\url{https://www.bing.com/translator}} is the company's flagship translation model. The architecture details and model size for this model are not publicly listed. The translations for each source target pair for this model were also  generated between 17 February 2024 to 19 February 2024 using their API.

\item 
\textbf{Seamless}:
Meta's SeamlessM4T (Massively Multilingual \& Multimodal Machine Translation) is a single model that supports speech to-speech, speech-to text, text-to-speech, text-to-text translation, and automatic speech recognition for up to 100 languages \citep{barrault2023seamlessm4t}. We use the SeamlessM4Tv1 variant with 2.3B parameters\footnote{\url{https://huggingface.co/facebook/seamless-m4t-large}}. This is the only multimodal model in our study.

\item 
\textbf{gold}:
Gold refers to the existing references in the ratings data we collected from various sources as described in the previous section. The internal data are human post-edited while the provenance of the references for the publicly available dataset are not clear. We assume these to be a mix of human translated and human post-edited references.

\item 
\textbf{oldX}:
OldX refers to an encoder-decoder architecture based transformer model as per the architecture described by \cite{mujadia2022ltrc}. The same architecture was extended to other language pairs and the resulting models used in the study. The \textit{oldX} models were intended as the translation baseline for all the language pairs.

\item 
\textbf{perturb}:
Perturb refers to perturbed translations. The perturbations were injected into the gold references and were of the following four types: \textit{Random Insertion, Random Substitution, Random Deletion, and Synonym Replacement}. The perturbation types were inspired by the method described in \cite{wei2019eda}. The \textit{Synonym Replacement} type required a lexical knowledge base. For English source, we utilized WordNet \citep{miller1995wordnet}, while for Hindi we used IndoWordNet \citep{bhattacharyya-2010-indowordnet}. Languages which were not supported by IndoWordNet skipped the \textit{Synonym Replacement} perturbation. For each sample, a perturbation type was randomly chosen and 20\% of the words (again randomly picked) in the target sentence were perturbed.
    
\item 
\textbf{perturbed\_multiple}:
Perturbed\textunderscore multiple refers to multiple perturbations applied to the gold reference. The perturbation types were the same as described above but in this case \textbf{two} types of perturbation were performed for each sample. The two perturbation types were chosen randomly from all possible permutations. The same constraints applied to the \textit{Synonym Replacement} perturbation based on whether the target language was supported in the IndoWordNet.

\end{itemize}

In all we had 10 distinct target hypothesis available to be paired with each source. The targets were diverse in quality and were representative of the architectures and engines currently popular in the field.

\subsubsection*{Target References}
As mentioned earlier, there was no special preprocessing required for the target gold references. Since our data came from existing parallel corpora, the domain labelling and sentence length bucketing was performed on the source and the paired target was simply picked as the representative gold reference.

\subsection{Task Planning and Execution}

\subsubsection*{Workbench}

\begin{figure}[ht]
  \centering
  \includegraphics[width=1\textwidth]{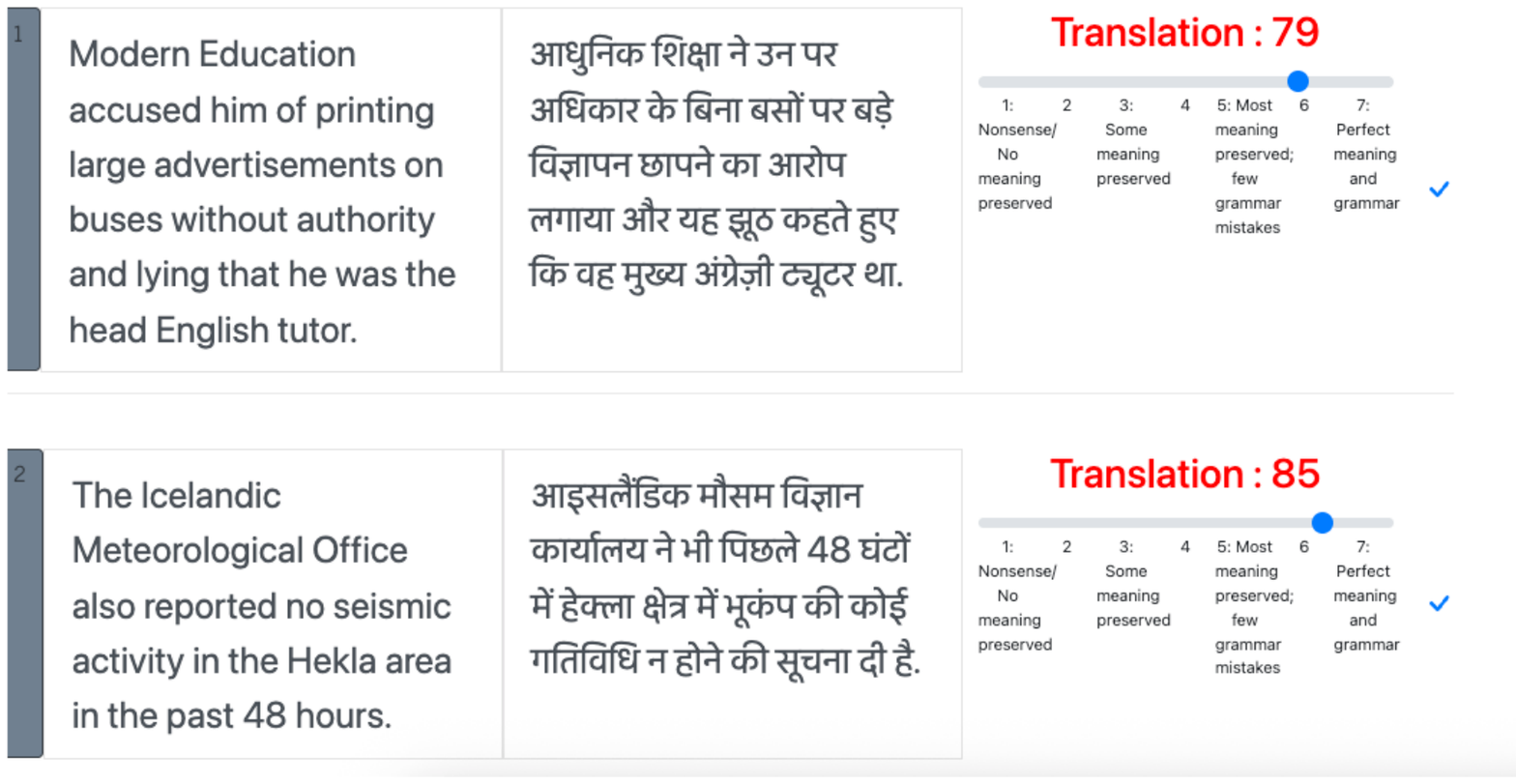}
  \caption{Posteditme, the cloud-based platform used for the rating task. An English source and a Hindi target are shown side by side along with a DA+SQM scale. Raters see the scores change upon moving and positioning the slider. Unlike WMT, there is no accompanying document context.} 
  \label{fig:posteditme}
\end{figure}

We utilized an in-house cloud-based post-editing workbench called \textit{Posteditme} that was extended to MT evaluation tasks.\footnote{\url{https://posteditme.in}} A screenshot from one of the practice runs depicting a scored English-Hindi pair is shown in Figure \ref{fig:posteditme}.

\subsubsection*{Rater Onboarding}
Once evaluators had been selected as per the methodology discussed in Section \ref{sec:data-raters}, their accounts were created on the \textit{Posteditme} platform where two sample rating tasks containing 10 segments each were uploaded for their chosen evaluation pair. The purpose of this initial exercise was to familiarize them with the platform and the rating task. In an online orientation meeting all participants were informed that they would be required to rate a minimum of 3000 items to qualify for payment. This was to reduce the chances of participants abandoning the task midway, for any ratings generated below these many items would have lowered the power of the statistical tests to be performed to filter consistent and inconsistent raters. However, this was not strictly enforced as there were not many drop-outs. Everyone was paid the going market rate.

Post completion of the practice tasks, the actual task files were uploaded into each rater's account.

\subsubsection*{Task Distribution}
\label{sec:data-planning-distribution}
In preparation for the ratings task, the pooled data were grouped by each language pair and domain. Our goal was to have at least 3000 unique source segments paired with 5 distinct translations. The translation hypotheses listed in Section \ref{sec:dataset-prep}, were first grouped into four types:

\begin{enumerate}[label=(\roman*), align=left]
\item \textbf{primary}: Bing, Google, Seamless, IndicTrans2, versionvN
\item \textbf{degraded}: oldx, perturbed, perturbed\_multiple
\item \textbf{LLM}: GPT-3.5
\item \textbf{human}: gold
\end{enumerate}

We then randomly sampled two translations from the \emph{primary} group and one each from the remaining groups to pair with each source. These pairings resulted in 15000 source-target items to be rated per pair. We strove to ensure that each item was rated by at least 5 raters. The data were grouped into task files, where each task file contained \textbf{44} segments to be rated which  also included \textbf{4} quality control items. We tried to enrol a minimum of 15 raters per language pair, however for some languages we were not able to meet that target.

Assuming 15 raters for each language pair, and based on some dry runs performed internally by the Authors themselves, we estimated that each rater may be able to rate 200-300 items per day. Based on this we set a task completion deadline of two weeks. Raters were free to attempt the tasks at any time during the day. The rating task's progress was constantly monitored and new files were uploaded into the raters' accounts based on their progress. Two meetings were held at the end of each week to communicate the overall status of the tasks per language pair and address any queries.

For some language pairs, we could not find the minimum number of raters (15) that we had estimated. We therefore kept advertising the task and any new raters who passed the Reading Comprehension Test were on-boarded in parallel even as the task progressed. While this resulted in an uneven number of ratings per pair, this was the only way to ensure that all pairs were covered. The entire ratings exercise conducted purely online took approximately four weeks from orientation to completion.

In the next section we present some descriptive statistics of the raw ratings data created via this entire exercise. 

\subsection{Raw Data Summary}

The entire exercise generated approximately \textbf{1 million} rating data points across 22 language pairs and involved 275 unique raters. 

\begin{figure}[ht]
  \centering
  \includegraphics[width=\textwidth]{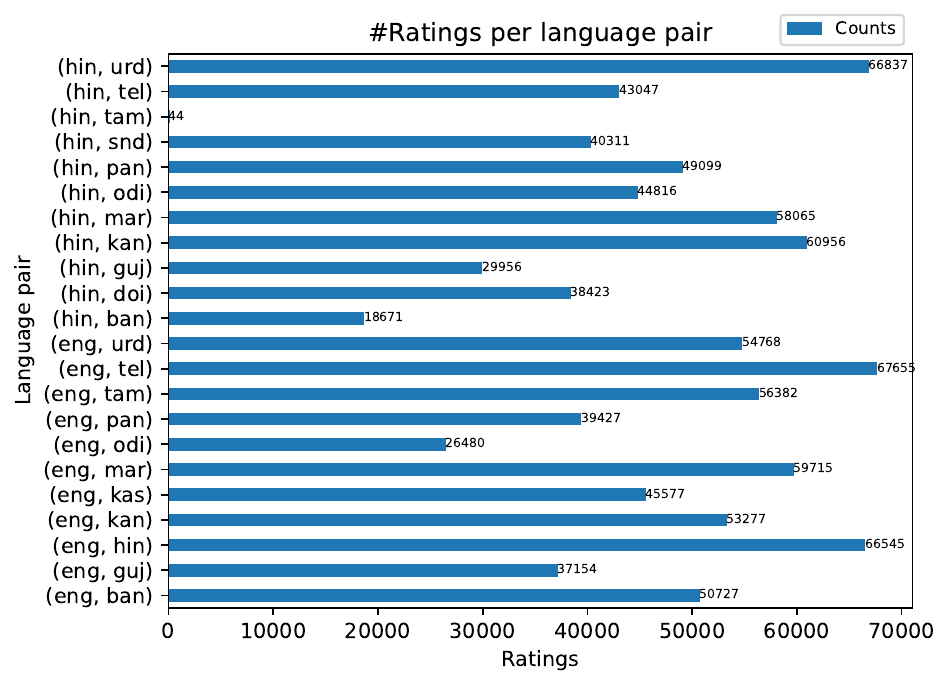}
  \caption{Number of ratings generated per language pair.}
  \label{fig:ratings_pair}
\end{figure}

Figure \ref{fig:ratings_pair} shows the number of rated items per language pair. \textit{hin-urd, eng-tel, and eng-hin} pairs contribute the highest numbers with more than 66,000 ratings each, while the \textit{hin-tam} pair could gather only 44 ratings. The variance in the number of ratings generated for each pair was dependent on the availability of bilingual raters for that translation direction (some, like \textit{hin-tam} harder to recruit) and the speed at which some  raters completed their tasks.

\begin{figure}[h]
  \centering
  \includegraphics[width=\textwidth]{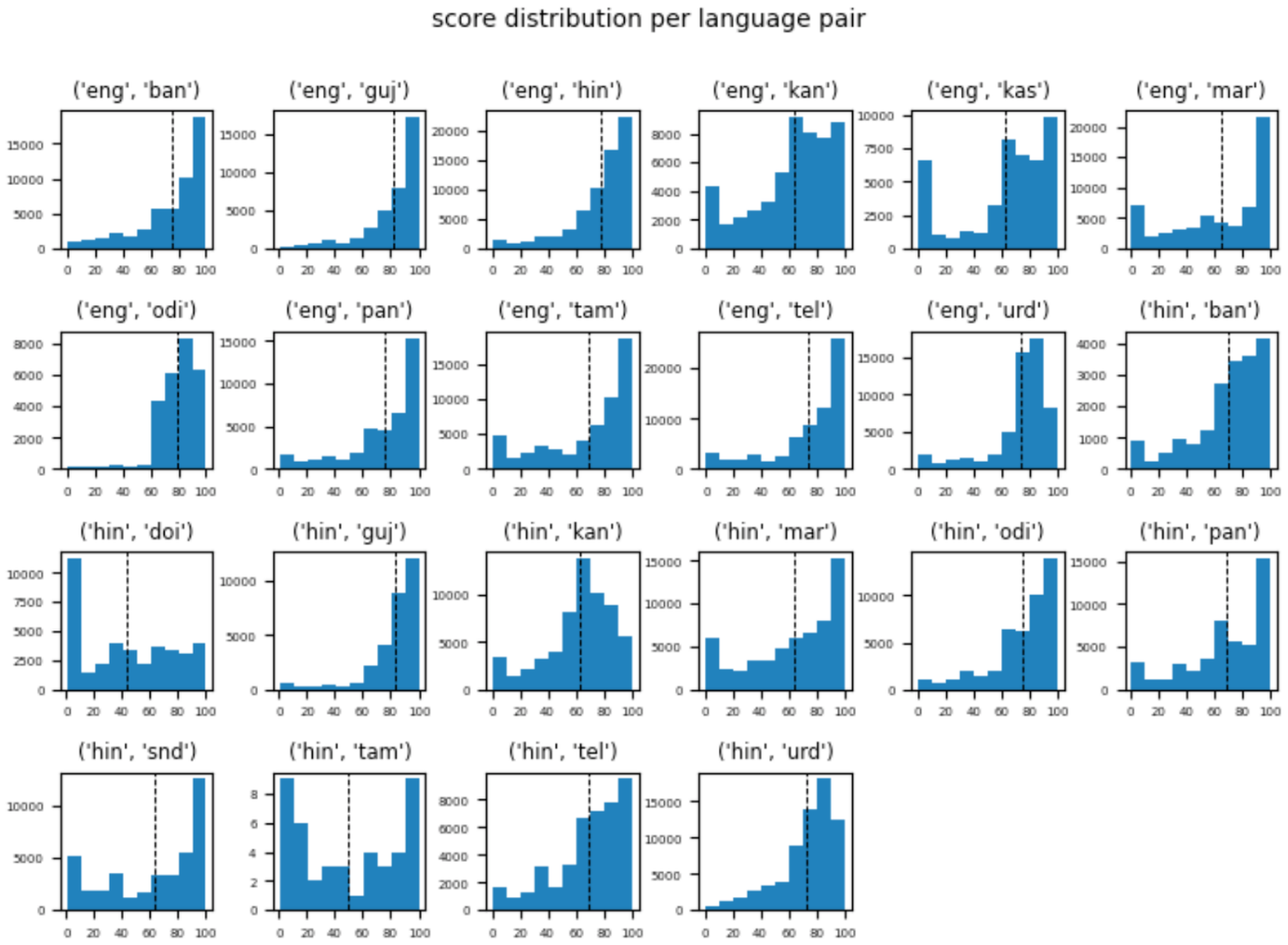}
  \caption{Score distribution per language pair on raw data. The shape of the distribution is an early indicator of average translation quality for each pair.}
  \label{fig:raw_distribution}
\end{figure}

Figure \ref{fig:raw_distribution} summarizers the distribution of the DA+SQM scores per language pair. The shape of the distribution is an early indicator of the overall translation quality available for that language pair. As expected, language pairs such as \textit{eng-hin} and \textit{eng-guj} with an established digital presence, tend to have left-skewed distributions, implying higher translation quality. Also noteworthy is that with \textit{Hindi} as source, the distribution shapes vary greatly depending on the target. 

Also see Figure \ref{fig:raters_pair} in Appendix \ref{app:details-raw-summary} for the number of raters that participated for each language pair. Figure \ref{fig:diversity_pair} from the same section shows each individual rater's contribution towards their language pairs total counts. Note that none of the language pairs, even the ones with low rating counts, show ratings dominated by only a few raters.  This rating diversity was controlled for at the time of task assignment and why all task files were not uploaded at once, rather released based on each rater's progress.

\section{Experimental Setup}
\label{sec:experiment}

\subsection{Data}
Starting with the raw data described in the previous section, we pre-process it to add new columns with additional derived information, filter it on quality control items to remove inconsistent raters, and normalize it further to arrive at our training dataset.

\subsubsection{Preprocessing}
We derive a number of additional columns from the raw data. For each source and target pair that had been rated, we add a reference column, which is the gold translation already available and in most cases also rated for each source sentence.

We also add columns counting the total number of ratings and total number of raters per item. All ratings for each unique item and the rater-ids for the item are also aggregated and listed. The mean of the DA+SQM ratings, the domain of the item, and the source length bucket are also augmented to the dataset. 

We then calculate the \textit{z-scores} of all the items across a language pair after grouping by raters following the standard practice from WMT shared tasks \citep{bojarFindings2018Conference2018}. We then derive the ratings and mean ratings columns again, but this time, with the standardized scores. 

In preparation of Quality Control based filtering, and for the purposes of planned ablation experiments, we also classify each rated target hypothesis's origin with a \textit{good}, \textit{bad} or \textit{neutral} label. We define \textit{versionvN, gold, MS-Bing, and Google} outputs to be \textit{good}, whereas \textit{perturb, oldX, perturbed-multiple}, the degraded outputs, are classified as \textit{bad}, and the remaining make up the \textit{neutral} category.

Owing to the uneven number of ratings that we ended up with, there were items for which we had missing references (since they had not yet been rated when the task ended). We discard all such items (approx. 3.4\% of the total data points) for which we could not locate a reference within the rated data.

\subsubsection{Quality Control Tests}

As previously mentioned in Section \ref{sec:data-planning-distribution}, in our setup each task file containing 44 items also included 4 quality control (QC) items. These items consisted of one each of a \textit{good} and a \textit{bad} translation hypothesis sampled randomly and also their repetitions within the same file. Grouping the data by raters and focusing only on the ratings of these quality control items, we set about filtering for two types of raters and their ratings from our data. For this we utilize two types of statistical significance tests described below.

\textbf{Consistency Test:} Consistent Raters are defined to be those raters who are consistent in their scoring of \textit{original} and \textit{repeated} items. In this case our null hypothesis $H_0$ states that the difference between the means of the rater's scores on the \textit{original} and \textit{repeated} items is $0$. For $\alpha < 0.5$, we reject $H_0$, which implies that these are the raters who scored the \textit{original} and \textit{repeated} items significantly differently and therefore are \textbf{NOT consistent} and should be filtered out. We use the related samples \textit{t-test} for this\footnote{\url{https://docs.scipy.org/doc/scipy/reference/generated/scipy.stats.ttest\_rel.html}}.
    
\textbf{Discernment Test:} Discerning Raters are defined as raters who can successfully discern quality differences between \textit{good} and \textit{bad} targets. Our null hypothesis $H_0$ states that the difference between the means of the rater's scores on the \textit{good} and \textit{bad} items is $0$. For $\alpha < 0.5$, we reject $H_0$ which means that these are the raters who scored the \textit{good} and \textit{bad} items significantly differently, and therefore are \textbf{discerning} and should NOT be filtered out. We use the independent samples \textit{t-test} for this\footnote{\url{https://docs.scipy.org/doc/scipy/reference/generated/scipy.stats.ttest\_ind.html}}.

We apply the consistency test in turn first to \textit{all} QC items, then only the \textit{good} QC items, and finally the \textit{bad} QC items, but we filter using the results of the \textit{bad} (original and repeated) items only. This is in line with WMT and makes sense because \textit{bad} or degraded items should ideally stand out more to consistent raters \citep{bojar2016findings}. 

For the discernment test we filter only on the scores of the original \textit{good} and \textit{bad} items and ignore the repeats. Note that while filtering for non-discerning raters, we do not filter raters if they belong to one of Kashmiri, Sindhi, or Dogri, owing to the fact that there would be very little difference between the \textit{good} and \textit{bad} targets for these languages as they are quite low-resourced; this is also borne out by the raw statistics of the ratings data generated for these languages.

\textbf{Filtering:}
On applying these tests, we uncover 8 raters who fail the consistency test and 84 raters who fail the discernment test. Note that this includes raters who had abandoned the tasks midway and did not generate enough ratings; in fact they make up the bulk of the raters who do not make the final cut. Only 21 out of 84 raters who failed the discernment test, contributed more than 3000 ratings (the threshold communicated to qualify for payment). We end up discarding $147030$ items, $14.6\%$ of the total rated items when we filter out non-discerning raters and $47615$ items, $4.7\%$ of the total rated items when we filter out inconsistent raters. Note that these numbers are far lower than those reported in WMT crowd-sourced rating tasks, where up to $60\%$ of the ratings end up being discarded \citep{kocmiFindings2022Conference2022}. 

We also discard any items that had only one rating when the task ended, about $1.3\%$ of the original data.

For this filtered data we recalculate column values that are affected by the change in volume of the data, such as number of ratings, number of raters, and most importantly, means per item. We then recompute the z-scores.

Recall that each unique pair was rated by up to $5$ raters. We now \textit{fold} the data, that is reduce it to list only unique source-target sentences along with their aggregated scores across all raters. This yields a final dataset of $221941$ items.

\subsubsection{Min-max Normalization}

The z-scores at first are unbounded scores centered at $0$. In order to keep the scores interpretable, we follow the custom \textit{min-max} scaling method described in \cite{guerreiro2023xcomet}. We set the data minimum to be the mean z-score of all the z-scores of those items, where the raw scores were $1$ across all raters. Similarly, we set the data maximum to be the mean z-score of all the z-scores of items where the raw scores were $100$ across all raters. We then apply min-max scaling on all the z-scores and clip all resulting values \textit{less than $0$} to $0$ and all values \textit{greater than $1$} to $1$. This is the normalized rating score on which we train all our models, except for the LLM-based models, which are trained on raw scores.

\subsubsection{COMTAIL Dataset}

We create \textit{train, dev, test} splits on this data in ratio of 0.9, 0.05, 0.05. We group the data by source and target language pairs and randomly sample for each split. Since the data vary greatly across languages pairs this ensures proportional representation in each split. 

Finally, we identify and filter out any source-target combinations from the dev and test splits that may have been seen in the training set. This ensures that the test set is as unbiased as possible.\footnote{source-target combinations may be duplicated across the train, dev, and test splits because different translation hypotheses can hypothetically result in the same translation. Although each item may have been scored differently.} Note that even a stricter filtering was possible to ensure that even the source remained unseen across train, dev and test splits, but this would have led to almost $50\%$ of the created data being filtered out. We therefore apply only the source-target filter with the implication that some source sentences from our test and dev split may also be found in the training data. Table \ref{tbl:train-dev-test} shows the statistics of the train, dev and test sets per language pair.

\begin{table}[ht]
\begin{minipage}{\textwidth}
\caption{COMTAIL Dataset. train, dev, test splits are shown per language pair along with aggregated totals over all language pairs.}
\label{tbl:train-dev-test}
\centering
\begin{tabular}{llrrr}
\toprule
srclng & tgtlng & train & dev & test \\
\midrule
eng & ban & 9823 & 484 & 481 \\
eng & guj & 8232 & 408 & 401 \\
eng & hin & 12164 & 567 & 571 \\
eng & kan & 11212 & 542 & 559 \\
eng & kas & 9773 & 501 & 491 \\
eng & mar & 11853 & 583 & 598 \\
eng & odi & 2023 & 99 & 103 \\
eng & pan & 7240 & 345 & 349 \\
eng & tam & 11719 & 543 & 538 \\
eng & tel & 12127 & 572 & 615 \\
eng & urd & 10639 & 533 & 525 \\
hin & ban & 5468 & 280 & 274 \\
hin & doi & 8726 & 443 & 451 \\
hin & guj & 7684 & 379 & 379 \\
hin & kan & 10555 & 520 & 525 \\
hin & mar & 11826 & 583 & 597 \\
hin & odi & 8950 & 458 & 456 \\
hin & pan & 10522 & 517 & 513 \\
hin & snd & 8868 & 451 & 448 \\
hin & tel & 8190 & 405 & 403 \\
hin & urd & 12143 & 602 & 610 \\
\midrule
all & all & 199737 & 9815 & 9887 \\
\bottomrule
\end{tabular}
\end{minipage}
\end{table}

The entire data described in Table \ref{tbl:train-dev-test} are the training, development, and test sets used in building all subsequent models, except for the ablated models, where different subsets of the same data are utilized and described separately.

\subsubsection{WMT Datasets}

We also utilize the following WMT datatsets to train models in combination with our dataset:

\paragraph{WMT-DA} This dataset contains all DA human annotations from previous WMT News Translation shared tasks held between 2017--2022 \citep{bojar2017findings,bojarFindings2018Conference2018,barraultFindings2019Conference2019,barraultFindings2020Conference2020,akhbardehFindings2021Conference2021,kocmiFindings2022Conference2022}.

\paragraph{WMT-MQM} This dataset contains MQM human annotations from previous WMT Metrics shared tasks and the MQM annotations from \cite{freitag2021experts}.

\paragraph{WMT-SQM} This dataset is from the WMT's General Translation task from 2022. This was the first time that the DA+SQM (Direct Assessment + Scalar Quality Metric) methodology was used \citep{kocmiFindings2022Conference2022}.

Instead of the custom min-max normalization method described earlier, we apply default min-max normalization and clipping on these datasets using \texttt{MinMaxScaler} from \texttt{scikit-learn}. WMT-DA and WMT-MQM datasets already contained z-cores from having been standardized for individual rater's scoring variations. However, WMT-SQM only had the raw DA+SQM score on a 0--100 scale and did not contain any rater information or provide standardized scores. We nevertheless normalized the raw scores with min-max normalization to bring all the dataset scores to the same scale.

Note that none of these datasets had existing training, development, and test splits. We create our own over the entire data, except for the WMT-DA dataset, where we subset the data by year and use the data from 2022 exclusively for development and test splits. We combine the train splits with the COMTAIL dataset to build additional combined models.

\begin{table}[h]
\caption{Overview of datasets, their combinations, annotation types, and volumes. Sizes are listed as per the derived training sets and their combinations. *Models are built only for the marked dataset names.}
\label{tbl:datasets-overview}
\centering
\begin{tabular*}{\textwidth}{@{\extracolsep{\fill}}l l l r} 
\hline
name & datasets combined & annotation type & size  \\
\hline
 comtail\textsuperscript{*} & comtail & da+sqm  &  199737 \\
 wmt-da & wmt-da & da & 1273930 \\
 wmt-mqm & wmt-mqm & mqm & 135313 \\
 wmt-sqm & wmt-sqm & da+sqm & 98645 \\
 comtail-wmt-da\textsuperscript{*} & {comtail, wmt-da} & da, da+sqm & 1473667 \\
 comtail-wmt-all\textsuperscript{*} & {comtail, wmt-da, wmt-mqm, wmt-sqm} & da, da+sqm, mqm & 1707623 \\
\hline
\end{tabular*}
\end{table}

Table \ref{tbl:datasets-overview} lists the sizes of the datasets individually and in combination. We only build models with the following datasets: comtail, comtail-wmt-da, and comtail-wmt-all.

We now proceed to model building and describe the architectures, parameters, and other associated details. We build both reference-based and reference-less models by training them from scratch, and where applicable also fine-tune an available base model.

\subsection{COMTAIL Models}
\paragraph{COMTAIL-DA}

\begin{figure}
\centering
\includegraphics[width=0.9\textwidth]{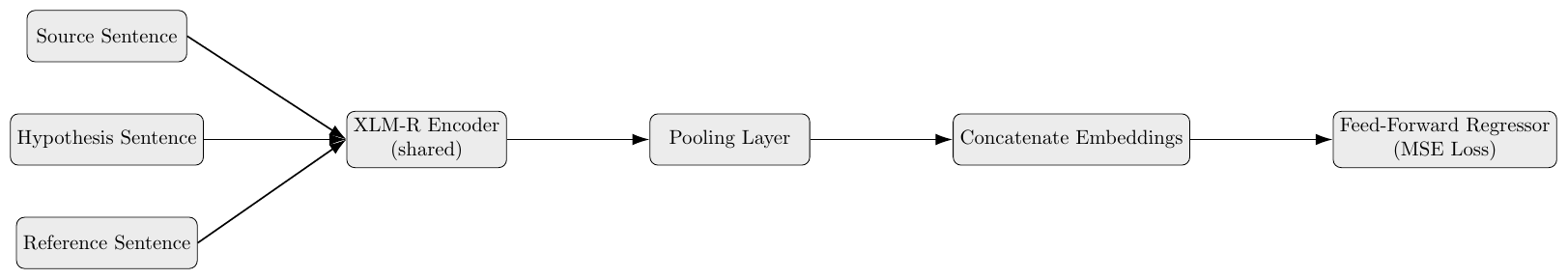}
\caption{COMET estimator model architecture. (Adapted from \cite{rei-etal-2020-comet}).} 
\label{fig:comet-architecture}
\end{figure}

We \textbf{train} a \textit{COMTAIL-DA} model from scratch on the COMET architecture as proposed in \cite{rei-etal-2020-comet}. COMET is described as a neural framework for training multilingual machine translation evaluation models. The COMET architecture is unique in that it incorporates the source in addition to the translation hypothesis and reference. We utilize the estimator model architecture where source, hypothesis, and reference are independently encoded using a pre-trained crosslingual encoder. We use XLM-RoBERTa as the encoder model \citep{conneau-etal-2020-unsupervised}. The resulting word-level embeddings are then passed through a pooling layer to create a sentence embedding for each segment. These embeddings are then again combined and concatenated into one single vector that is passed to a feed-forward regressor. The entire model is trained by minimizing the Mean Squared Error (MSE). See Figure \ref{fig:comet-architecture}.

We also \textbf{fine-tune} a COMTAIL-DA model using \textit{wmt22-comet-da} as the base model.\footnote{\url{https://huggingface.co/Unbabel/wmt22-comet-da}} We utlize the entire COMTAIL dataset from Table \ref{tbl:train-dev-test} for both training and fine-tuning.

We train or fine-tune all models for a maximum of $5$ epochs. See Table \ref{tab:hyperparams} for more details. 

\paragraph{COMTAIL-QE}
We \textbf{train} a reference-less QE model as per the architecture described in \cite{rei2021references}. The QE model training architecture is quite similar to the estimator architecture described previously for DA models, except that reference sentences are not used. The underlying encoder model (XLM-RoBERTa Large) remains unchanged.


However, to \textbf{fine-tune} the QE models, we use the wmt22-cometkiwi-da model as the base model, which is the flagship COMET QE model currently listed.\footnote{\url{https://github.com/Unbabel/COMET?tab=readme-ov-file\#comet-models}} It is based on a multitask architecture that tries to estimate word labels as well as a quality score \citep{rei2022cometkiwi}. However, in our setting as we are not estimating word-level tags, we fine-tune only for sentence-level scores.


\paragraph{Combination Models}

Following the same architectures, we also train and fine-tune both DA and QE models for the \textit{comtail-wmt-da} and \textit{comtail-wmt-all} combined datasets. We propose to contrast the trained and fine-tuned models built with this additional data with the COMTAIL models built solely on the dataset we created.

\subsection{Llama-based Models}
\paragraph{COMTAIL-llama-DA}
We fine-tune a \textit{LLaMA 3.1-8B} base model on the COMTAIL dataset \citep{grattafiori2024llama3herdmodels}. We use Meta’s \textit{llama-recipes} library, applying Parameter-Efficient Fine-Tuning (PEFT) with LoRA adapters and int8 quantization on a single \textit{RTX 4090} GPU \citep{llama-recipes, hu2021lora}.

Following previous efforts in this direction, we fine-tune the model first on the raw mean scores from the original DA+SQM scale of 0--100 \citep{freitag2022gemba}. We then also fine-tune it on the normalized mean z-scores between 0--1. We hypothesize that the raw scores would provide a better training signal during fine-tuning of these models when they are used as text generators of these 'scores' instead of being trained as regressors.

\begin{figure}[h]
  \centering
  \includegraphics[width=0.9\textwidth]{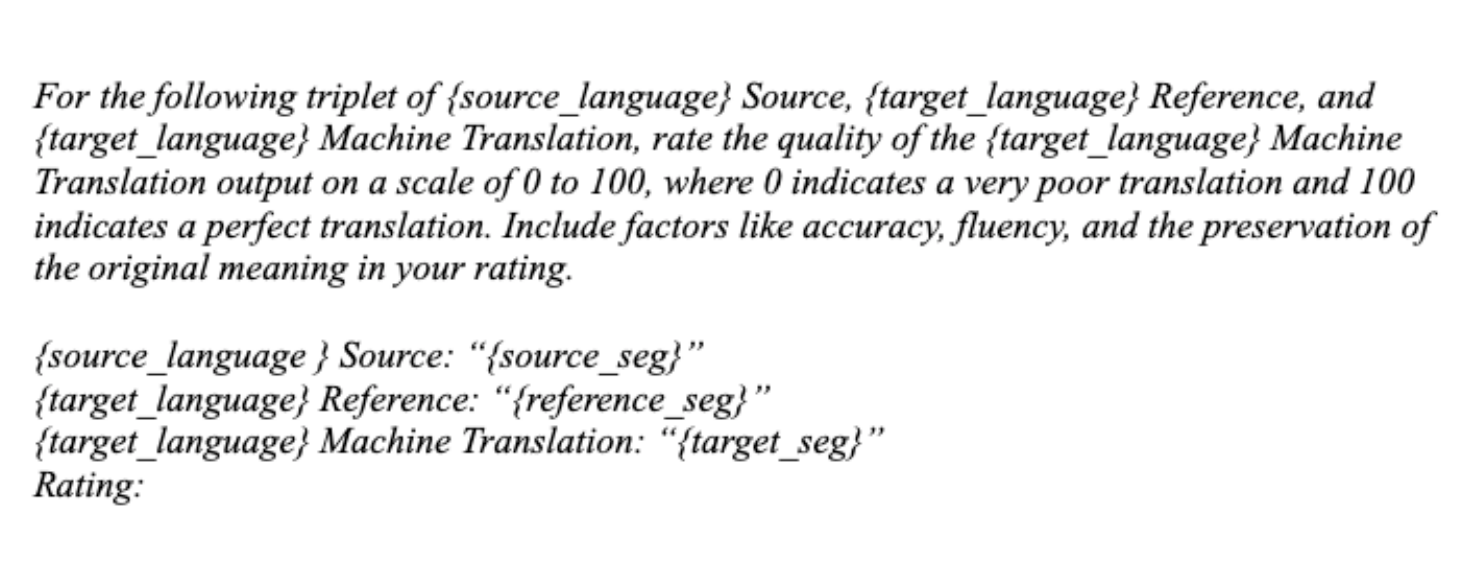}
  \caption{Prompt template used for the LLaMA 3.1-8B base model on COMTAIL dataset \textbf{with} references.}
  \label{fig:da-prompt}
\end{figure}

Figure \ref{fig:da-prompt} shows the prompt used for finetuning with the entire COMTAIL dataset. A triplet of source, hypothesis, and target are supplied along with a mean DA+SQM assessment score on a scale of 0--100. A separate model is also trained in a similar manner using the existing scaled and clipped z-scores in the range of 0--1.

\begin{figure}[h]
  \centering
  \includegraphics[width=0.9\textwidth]{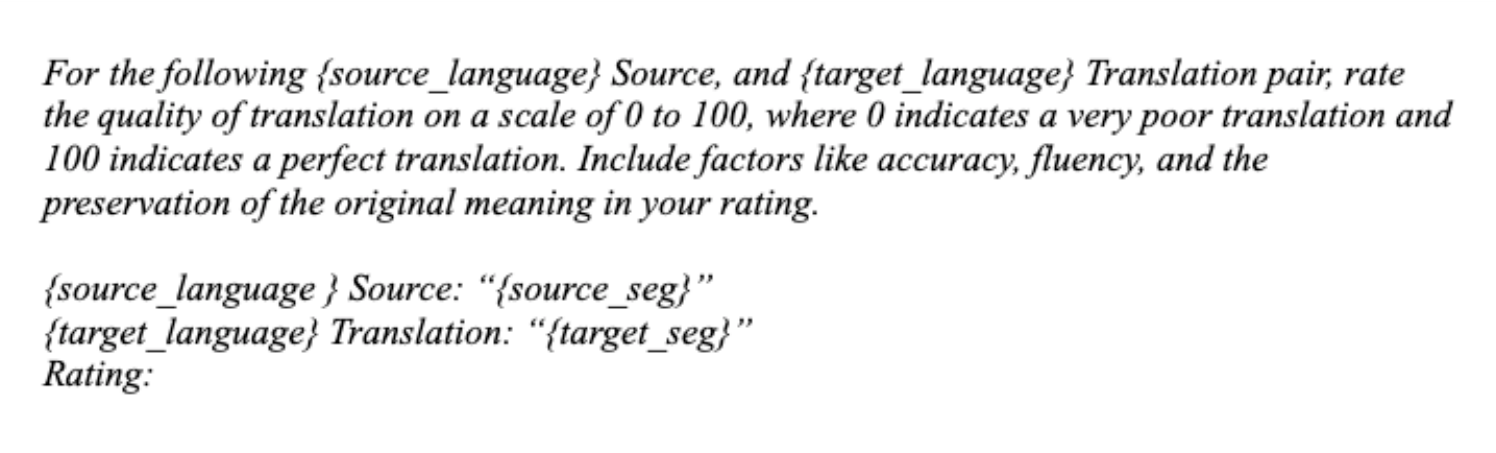}
  \caption{Prompt template used to fine-tune the LLaMA 3.1-8B base model on COMTAIL dataset \textbf{without} references.}
  \label{fig:qe-prompt}
\end{figure}

\paragraph{COMTAIL-llama-QE}
We also fine-tune a QE reference-less model. Figure \ref{fig:qe-prompt} shows the prompt used for finetuning with the entire COMTAIL dataset. A source and a hypothesis are supplied along with a mean DA+SQM assessment score on a scale of 0--100. A separate model is also trained in a similar manner using the existing scaled and clipped z-scores in the range of 0--1.

\subsection{Baseline Models}
We compare against a number of publicly available models.

\paragraph{COMET-DA}
\textit{comet-wmt-22-da} is currently the flagship DA model from the COMET family of models.\footnote{\url{https://github.com/Unbabel/COMET?tab=readme-ov-file\#comet-models}} Based on the COMET estimator architecture, it employs a reference-based regression approach and is built upon the XLM-R encoder. It is trained on direct assessments from WMT17 to WMT20 and provides scores ranging from 0 to 1 \citep{reiCOMET22UnbabelIST20222022}.

\paragraph{COMET-QE}
\textit{comet-kiwi-wmt22-da} is the flagship quality estimation COMET model \citep{rei2022cometkiwi}. It employs a regression approach and is built on top of InfoXLM \citep{chi-etal-2021-infoxlm}. Like its DA counterpart, it is also trained using direct assessments from WMT17 to WMT20, as well as direct assessments from the MLQE-PE corpus. It generates scores ranging from 0 to 1.

\paragraph{Explainable COMET}
\textit{comet-xl} model is from the eXplainable COMET (XCOMET) variants. It is trained to identify error spans and assign a final quality score, resulting in an explainable neural metric. We utilize only the quality scores from this model and use the XL variant with 3.5 billion parameters \citep{guerreiro2023xcomet}.

\paragraph{IndicCOMET-DA}
The IndicCOMET-DA model that we label as \textit{da-ai4b} is a  COMET-based model trained on the Indic-MT Eval Dataset \citep{saibIndicMTEvalDataset2023}. The Indic-MT Eval Dataset contains MQM annotations for about 5 Indian languages consisting of 7000 annotated sentences. The languages covered are: Tamil (tam), Gujarati (guj), Hindi (him), Marathi (mar), and Malayalam (mal). IndicCOMET has a DA and an MQM variant, however, in order to make a fair comparison, we compare against the IndicCOMET-DA model, trained by initializing with the COMET-DA model. 

\subsection{String-based Metrics}
In addition to the models described above we also compare the flagship model's results against the following lexical overlap-based metrics. 
\textbf{ 
 Bilingual Evaluation Understudy} (BLEU) is a corpus level metric and has been the most widely used metric for MT evaluation \citep{papineni2002bleu}. Although, typically, BLEU is calculated at the corpus level, we use an implementation of sentence level BLEU in order to compare scores with all the other sentence level metrics in our study. \textbf{Translation Edit Rate} (TER) is a sentence level metric that quantifies the edit operations needed on an MT hypothesis to turn it into a given reference \citep{snover2006study}. It is the only metric in our experiments where a \textit{lower} score implies a better MT output. \textbf{
character n-gram F-score} (CHRF) is a sentence level metric that uses character n-grams as its countable units \cite{popovic2015chrf}. We utilize the recommended \begin{math}\beta\end{math} value of 2 \citep{popovic2016chrf}. For all three string-based metrics we use the ScareBLEU implementation \citep{post2018call}.

\subsection{Ablation Setup}

The effect of data volume, domain, quality, and the linguistic relatedness of the languages constituting ratings data on models built with such data is to the best of our knowledge, currently understudied. We set up a number of ablation experiments which we hope would shed light on some of the following questions:

\begin{enumerate}
    \item What is the minimum volume of ratings data required to extend a neural evaluation metric to new pairs?
    \item How robust are neural metrics? How do metrics trained on specific domain data perform on unseen domains?
    \item Do hypothesis quality variations have an effect on the metric?
    \item What role does linguistic relatedness play?
    \begin{enumerate}
        \item Does the source not being part of model training data affect the target inference?
        \item How do language families behave on metric models built without their data?
        \item How do language \textit{sub}-families behave on metric models built without their data?
    \end{enumerate}
\end{enumerate}

To create our ablation data, we start with the full dataset from Table \ref{tbl:train-dev-test} and arrive at the train, dev, and test sets for each ablation experiment by filtering on the full dataset.

In all cases, where applicable we report the results on an ablated test set, else results are reported on the full test set.

\begin{table}[h]
\centering
\begin{minipage}[t]{0.45\textwidth}
\caption{Volume Ablation data}
\label{tbl:abl-volume-data}
\begin{tabular*}{\textwidth}{@{\extracolsep{\fill}}l l l l l} 
\toprule
volume & train & dev & test \\
\midrule
10\% & 19974 & 9815 & 9887 \\
25\% & 49934 & 9815 & 9887 \\
50\% & 99868 & 9815 & 9887 \\
75\% & 149803 & 9815 & 9887 \\
100\% & 199737 & 9815 & 9887 \\
\bottomrule
\end{tabular*}
\end{minipage}
\hfill
\begin{minipage}[t]{0.45\textwidth}
\caption{Quality Ablation data}
\label{tbl:abl-quality-data}
\centering
\begin{tabular*}{\textwidth}{@{\extracolsep{\fill}}l l l l} 
\toprule
quality & train & dev & test \\
\midrule
bad	& 44067 & 2214 & 2281 \\
good & 86971 & 4210 & 4107 \\
neutral	& 68699 & 3391 & 3499 \\
\bottomrule
\end{tabular*}
\end{minipage}

\vspace{0.5cm} 

\begin{minipage}[t]{0.45\textwidth}
\caption{Domain Ablation data}
\label{tbl:abl-domain-data}
\begin{tabular*}{\textwidth}{@{\extracolsep{\fill}}l l l l} 
\toprule
domain & train & dev & test \\
\midrule
governance & 65477 & 3245 &	3247 \\
health & 52757 & 2578 & 2630 \\
\bottomrule
\end{tabular*}
\end{minipage}
\end{table}

\subsubsection{Volume}

We start by investigating the data volume question. Did we create enough data to improve metric performance on Indian Languages within the scope of our study, or could we have achieved comparable performance with smaller amounts of data?

From the full training set we randomly sample data in the following proportions: 10\%, 25\%, 50\%, 75\%, and 100\% of the total data. We train/fine-tune models for each volume proportion and report the results on the \textit{full} test set. Table \ref{tbl:abl-volume-data} lists the volume ablated dataset.

\subsubsection{Quality}
As discussed previously we were careful to include translations from a diversity of engines and of varying quality. Does this contribute to metric robustness? Recall that have the origin of each candidate translation which has already been classified as \textit{good}, \textit{bad}, \textit{neutral}. We ablate and train models with each quality type. Recall that \textit{versionvN, gold, MS-Bing, Google} are classified as good, whereas \textit{perturb, oldX, perturbed-multiple} are classified as bad, and the remaining make up the neutral category. Table \ref{tbl:abl-quality-data} lists the quality ablated dataset.


\subsubsection{Domain}
There has been some recent work on how domain changes affect COMET like evaluation metrics, and whether they can be reliably applied to data from unseen domains \citep{zouharFineTunedMachineTranslation2024, zouhar2024pitfalls}.

We hope to add our findings to this line of enquiry. As previously mentioned, we infer three coarse domains \textit{general}, \textit{governance}, and \textit{health} based on our data sources. While the provenance of governance and health data sources can be easily established, the general domain data come from external sources, therefore we keep our ablations to only the governance and health domains. Table \ref{tbl:abl-domain-data} lists the domain ablated dataset.


\subsubsection{Language}

\begin{table}[h]
\centering
\begin{minipage}{0.45\textwidth}
\centering
\caption{Language (x-il) Ablation data}
\label{tbl:abl-xil-data}
\begin{tabular*}{\textwidth}{@{\extracolsep{\fill}}l l l l r} 
\toprule
language (x-il) & train & dev & test \\
\midrule
en-il	& 106805 & 5177 & 5231 \\
hi-il & 92932 & 4638 & 4656 \\
\bottomrule
\end{tabular*}
\end{minipage}
\hfill
\begin{minipage}{0.45\textwidth}
\centering
\caption{Language (family) Ablation data}
\label{tbl:abl-family-data}
\begin{tabular*}{\textwidth}{@{\extracolsep{\fill}}l l l l r}
\toprule
language (family) & train & dev & test \\
\midrule
dravidian & 53803 & 2582 & 2640 \\
indo-aryan & 145934 & 7233 & 7247 \\
\bottomrule
\end{tabular*}
\end{minipage}

\vspace{0.5cm} 

\begin{minipage}{0.55\textwidth}
\centering
\caption{Language (subfamily) Ablation data}
\label{tbl:abl-subfamily-data}
\begin{tabular*}{\textwidth}{@{\extracolsep{\fill}}l l l l r} 
\toprule
language (subfamily) & train & dev & test \\
\midrule
dv-south & 33486 & 1605 & 1622 \\
dv-south-central & 20317 & 977 & 1018 \\
ia-central & 34946 & 1702 & 1706 \\
ia-dardic & 9773 & 501 & 491 \\
ia-east & 26264 & 1321 & 1314 \\
ia-north & 8726 & 443 & 451 \\
ia-north-west & 26630 & 1313 & 1310 \\
ia-south & 23679 & 1166 & 1195 \\
ia-west & 15916 & 787 & 780 \\
\bottomrule
\end{tabular*}
\end{minipage}

\end{table}

The Indian subcontinent hosts four major language families: Indo-Aryan, Dravidian, Munda, and Tibeto-Burmese \citep{Masica1976}. The linguistic features of the subcontinent's languages have been studied and mapped to establish it as one \textit{linguistic area}, exhibiting traits of their respective families but also marked by diffusion across genetic relationships \citep{emeneauIndiaLingusticArea1956}. Motivated by these insights, we propose a series of ablation experiments to help uncover the behaviour of neural models when built using different subsets of data. We are particularly interested in the following:

\paragraph{x-il}
Does the source direction not being part of the trained model's ratings data affect correlations? We ablate the data and train models in english to indian-languages \textit{(en-il)} and hindi to indian-languages  \textit{(hi-il)} directions. We inference the test sets after similar filtering. If our goal is to evaluate \textit{hi-il} data should we create data in this direction or will \textit{en-il} directions suffice? Table \ref{tbl:abl-xil-data} shows the ablated volumes.


\paragraph{family}
Do data from two distinct language groups (Indo-Aryan and Dravidian) help improve each other's correlations? We inference the test sets after similar filtering against models built exclusively on data from the two families.

Recall, that this is how the Indian languages used in this study group under the two families:
\begin{enumerate}
    \item indo-aryan = hin, ban, guj, kas, mar, odi, pan, urd, snd, doi
    \item dravidian = kan, tel, tam
\end{enumerate}

Table \ref{tbl:abl-family-data} lists ablated volumes for the two families.


\paragraph{subfamily}
Within the Indo-Aryan and Dravidian language families, how do subfamily groups fare on correlations? We investigate the following groupings:

Based on \cite{masica1991indo}, the Indo-Aryan languages in the dataset can be grouped as follows: ia-north: Dogri (doi); ia-north-western: Punjabi (pan), Sindhi (snd); ia-south: Marathi (mar); ia-east: Bengali (ban), Odia (odi); ia-western: Gujarati (guj); ia-central: Hindi (hin), Urdu (urd); ia-dardic: Kashmiri (kas).

And following \cite{krishnamurti2003dravidian}, we group the Dravidian languages from our dataset into: dv-south-central: Telugu (tel); dv-south: Tamil (tam), Kannada (kan). Table \ref{tbl:abl-subfamily-data} has the volume breakdown for each subgroup within the two families.

\paragraph{Hindi-Urdu}
The Hindi-Urdu 'controversy' is an enduring one. Are these the same language with two scripts with minor structural differences in their high registers \citep{masica1991indo, king1994one, schmidt2004urdu}, or do the highly evolved literary cultures around them mark them as two distinct languages \citep{faruqi2001early}?

We ablate Hindi and Urdu data from our datasets and put their respective models to test on each other's similarly ablated test sets.



\subsection{Challenge Set Evaluation}

There are a number of recent challenge sets that seek to evaluate machine translation quality on various linguistic and semantic categories \citep{isabelle2017challenge, avramidis2022linguistically, alves-etal-2022-robust,karpinskaDEMETRDiagnosingEvaluation2022, amrheinACESTranslationAccuracy2022a, mogheMachineTranslationMeta2024}. 

The ACES challenge set also represents this recent turn in machine translation evaluation and focuses on \textit{Accuracy} related translation phenomena linked to the MQM ontology. It consists of 36,476 examples covering 146 language pairs and represents challenges from 68 phenomena \citep{amrheinACESTranslationAccuracy2022a}. We seek to evaluate our metric against any pair that contains at least one Indian language in the challenge set; this yields us the following pairs: en–hi, en–mr, en–ta, en–ur, hi–en, mr–en, ta–en, ur–en, out of which the en-ta direction is discarded because it contains only 1 usable example.

Each item representing a linguistic phenomenon in the challenge set has a source, two translations and a reference. The goal is for our metric to correctly discriminate between the good and bad translation. Correlations are then calculated against the gold label and reported by grouping first on language pairs and then by phenomena.

\section{Results and Discussion}
\label{sec:results}

\subsection{COMTAIL Models}

\subsubsection*{DA}
Let us first look at the results of the reference-based models trained with the DA+SQM ratings. For the sake of brevity, we refer to them as DA models. 

Table \ref{tbl:comtail-da} lists Kendall’s Tau ($\tau$) correlations obtained for each language pair on the COMTAIL dataset. For each row, first the source and target languages are shown followed by the correlations obtained for the three surface metrics: blue, ter, chrf2. Correlation scores from baseline public models are then shown, the models being: COMET-DA (comet-wmt22-da), Explainable COMET (comet-xl), and IndicCOMET-DA (da-ai4b). These are followed by models obtained by fine-tuning a base COMET-DA model with the COMTAIL dataset and on a mixture of COMTAIL and WMT datasets as listed previously in Table \ref{tbl:datasets-overview}. The prefix \textit{ft-} marks these models. Finally, we show the results of the models trained from scratch,oto again utilizing the same three datasets mentioned earlier. These contain the prefix \textit{tr-} in their name.

All correlations are averaged across the 21 language pairs and a simple average is shown followed by a weighted average based on the contribution of each language pair to the entire test dataset. This was deemed necessary since the test counts per pair were unequal. We base all our analysis and discussion on the weighted average scores.

Looking at Table \ref{tbl:comtail-da} a few trends are noticeable. There is a large gap between the average surface metrics correlation scores and the public neural metric scores (from $0.29$ for BLEU to $0.44$ for COMET-DA, a jump of about $52\%$). This is not surprising given that embeddings-based metrics first outperformed surface metrics back in 2017 itself \citep{bojar-etal-2017-results}.

Both fine-tuning and training with the COMTAIL datatset imrpoves the correlations further to an average value of $0.53$, which translates to a $20\%$ increase over the current publicly available state-of the-art-models including the IndicCOMET-DA model fine-tuned with Indian language data. There seems to be very little difference in scores when fine-tuning a base model compared to training a model from scratch with different data combinations. Comparing the trained COMTAIL model (\textit{tr-comtail-ctl}) with the fine-tuned COMTAIL models (ft-comtail-ctl), we see very little difference in scores on average and even across language pairs. This seems to imply that the non-Indian languages that make up the target direction of the COMET model/dataset do not make a substantial contribution to Indian language evaluation directions.

The best performing model with a correlation value of \textbf{$0.53$} turns out to be the model fine-tuned using a base comet model utilziing the combined COMTAIL and WMT-DA datasets. 13 out of the 21 language pairs evaluated achieve the best correlations for this model. Amongst the models, the drop in correlation values for the \textit{eng-kas} and \textit{hin-doi} pairs for the larger explainable comet (comet-xl) model is surprising. We think, it is indicative of the impact on extremely low resource languages as model size increases. Another puzzling case is that of the \textit{eng-urd} and \textit{hin-guj} pairs, which show very little improvement when trained or fine-tuned with the COMTAIL dataset. This could point to noisy ratings despite QC-based filtering or imply data quality issues in the base pre-trained models itself, both of which call for further investigation.

\begin{table}[h]
\caption{DA Models: Kendall’s Tau ($\tau$) correlations obtained for each language pair on the COMTAIL dataset. ${*}$ marked row lists the simple average of the correlations across all language pairs that make up the test set. ${\dag}$ marked row lists the weighted average of the correlations as per each language pair's contribution to the total test set.}
\label{tbl:comtail-da}
\footnotesize
\centering
    \begin{tabular*}{\textwidth}{@{\extracolsep{\fill}}p{0.03\textwidth}p{0.03\textwidth}p{0.03\textwidth}p{0.03\textwidth}p{0.05\textwidth}p{0.03\textwidth}p{0.03\textwidth}p{0.03\textwidth}p{0.03\textwidth}p{0.03\textwidth}p{0.03\textwidth}p{0.03\textwidth}p{0.03\textwidth}p{0.03\textwidth}p{0.03\textwidth}p{0.03\textwidth}}
    \toprule
    \rotatebox{90}{\scriptsize src-lng} &
    \rotatebox{90}{\scriptsize tgt-lng} &
    \rotatebox{90}{\scriptsize count} &
    \rotatebox{90}{\scriptsize bleu} &
    \rotatebox{90}{\scriptsize ter} &
    \rotatebox{90}{\scriptsize chrf2} &
    \rotatebox{90}{\scriptsize comet-wmt22-da} &
    \rotatebox{90}{\scriptsize comet-xl} &
    \rotatebox{90}{\scriptsize da-ai4b} &
    \rotatebox{90}{\scriptsize ft-comtail-ctl} &
    \rotatebox{90}{\scriptsize ft-da-ctl-wmt-all} &
    \rotatebox{90}{\scriptsize ft-da-ctl-wmt-da} &
    \rotatebox{90}{\scriptsize tr-comtail-ctl} &
    \rotatebox{90}{\scriptsize tr-da-ctl-wmt-all} &
    \rotatebox{90}{\scriptsize tr-da-ctl-wmt-da} \\
    \midrule
    eng & ban & 481 & 0.28 & -0.31 & 0.26 & 0.54 & 0.53 & 0.46 & \textbf{0.58} & 0.56 & \textbf{0.58} & \textbf{0.58} & 0.57 & \textbf{0.58} \\
    eng & guj & 401 & 0.19 & -0.23 & 0.17 & 0.33 & 0.34 & 0.29 & 0.36 & 0.37 & \textbf{0.38} & 0.36 & 0.37 & \textbf{0.38} \\
    eng & hin & 571 & 0.20 & -0.23 & 0.15 & 0.47 & 0.53 & 0.48 & 0.55 & \textbf{0.58} & 0.57 & 0.56 & \textbf{0.58} & \textbf{0.58} \\
    eng & kan & 559 & 0.34 & -0.38 & 0.31 & 0.52 & 0.51 & 0.49 & 0.59 & 0.58 & 0.60 & 0.59 & \textbf{0.60} & 0.59 \\
    eng & kas & 491 & 0.36 & -0.31 & 0.37 & 0.39 & 0.06 & 0.29 & 0.49 & 0.52 & 0.50 & 0.49 & \textbf{0.54} & 0.51 \\
    eng & mar & 598 & 0.37 & -0.41 & 0.34 & 0.56 & 0.52 & 0.52 & \textbf{0.62} & \textbf{0.62} & \textbf{0.62} & 0.61 & 0.61 & 0.61 \\
    eng & odi & 103 & 0.29 & -0.35 & 0.30 & 0.43 & 0.43 & 0.44 & \textbf{0.48} & \textbf{0.48} & 0.45 & \textbf{0.48} & 0.46 & 0.46 \\
    eng & pan & 349 & 0.33 & -0.34 & 0.33 & 0.44 & 0.48 & 0.41 & 0.50 & \textbf{0.53} & 0.52 & 0.51 & 0.52 & 0.52 \\
    eng & tam & 538 & 0.32 & -0.31 & 0.33 & 0.51 & 0.48 & 0.46 & 0.57 & 0.58 & \textbf{0.59} & 0.58 & 0.58 & 0.57 \\
    eng & tel & 615 & 0.35 & -0.37 & 0.34 & 0.52 & 0.52 & 0.51 & 0.58 & 0.57 & \textbf{0.60} & 0.58 & 0.59 & 0.59 \\
    eng & urd & 525 & 0.32 & -0.31 & 0.31 & 0.40 & 0.39 & 0.38 & 0.41 & 0.40 & 0.41 & \textbf{0.42} & \textbf{0.42} & 0.41 \\
    hin & ban & 274 & 0.08 & -0.09 & 0.07 & 0.32 & 0.32 & 0.29 & 0.42 & 0.43 & 0.44 & 0.44 & \textbf{0.46} & 0.44 \\
    hin & doi & 451 & 0.54 & -0.52 & 0.52 & 0.50 & 0.10 & 0.42 & 0.66 & 0.62 & \textbf{0.67} & 0.64 & 0.66 & 0.64 \\
    hin & guj & 379 & 0.24 & -0.25 & 0.19 & 0.30 & 0.29 & 0.25 & 0.30 & 0.30 & \textbf{0.32} & 0.30 & \textbf{0.32} & 0.31 \\
    hin & kan & 525 & 0.31 & -0.35 & 0.28 & 0.49 & 0.46 & 0.47 & \textbf{0.55} & \textbf{0.55} & 0.54 & \textbf{0.55} & 0.54 & 0.53 \\
    hin & mar & 597 & 0.17 & -0.18 & 0.16 & 0.43 & 0.45 & 0.43 & 0.58 & 0.59 & \textbf{0.63} & 0.59 & 0.58 & 0.60 \\
    hin & odi & 456 & 0.21 & -0.22 & 0.20 & 0.27 & 0.26 & 0.26 & 0.31 & \textbf{0.33} & \textbf{0.33} & 0.32 & \textbf{0.33} & 0.32 \\
    hin & pan & 513 & 0.35 & -0.36 & 0.32 & 0.51 & 0.50 & 0.46 & \textbf{0.55} & \textbf{0.55} & \textbf{0.55} & \textbf{0.55} & 0.54 & 0.54 \\
    hin & snd & 448 & 0.27 & -0.29 & 0.27 & 0.39 & 0.41 & 0.39 & \textbf{0.55} & 0.53 & \textbf{0.55} & \textbf{0.55} & 0.53 & 0.53 \\
    hin & tel & 403 & 0.29 & -0.31 & 0.28 & 0.41 & 0.38 & 0.40 & 0.45 & \textbf{0.46} & \textbf{0.46} & \textbf{0.46} & 0.45 & 0.43 \\
    hin & urd & 610 & 0.17 & -0.17 & 0.15 & 0.35 & 0.45 & 0.37 & 0.47 & 0.49 & \textbf{0.50} & 0.47 & \textbf{0.50} & 0.49 \\
    \midrule
    all$^{*}$ & all & 9887 & 0.28 & -0.30 & 0.27 & 0.43 & 0.40 & 0.40 & 0.50 & 0.51 & 0.51 & 0.51 & 0.51 & 0.51 \\
    all$^{\dag}$ & all & 9887 & 0.29 & -0.30 & 0.27 & 0.44 & 0.41 & 0.41 & 0.51 & 0.52 & \textbf{0.53} & 0.52 & 0.52 & 0.52 \\
    \bottomrule
    \end{tabular*}
\end{table}

\subsubsection*{QE}
We next compare the reference-less models built with COMTAIL and WMT data combinations. Having established that the surface metrics have weak correlations with the DA+SQM ratings, we do not reproduce correlation values for them any further. The public QE model we compare against is COMET-QE's current SOTA model, \textit{comet-kiwi-wmt22-da}.

\begin{table}[h]
\begin{minipage}{\textwidth}
\small
\caption{QE Models: Kendall’s Tau ($\tau$) correlations obtained for each language pair on the COMTAIL test set. ${*}$ marked row lists the simple average of the correlations across all language pairs that make up the test set. ${\dag}$ marked row lists the weighted average of the correlations as per each language pair's contribution to the total test set.}
\label{tbl:comtail-qe}
\centering
\begin{tabular*}{\textwidth}{@{\extracolsep{\fill}}cccccccccc}

\toprule

\rotatebox{90}{\scriptsize srclng} &
\rotatebox{90}{\scriptsize tgtlng} &
\rotatebox{90}{\scriptsize count} &
\rotatebox{90}{\scriptsize comet-kiwi-wmt22-da} &
\rotatebox{90}{\scriptsize ft-qe-ctl} &
\rotatebox{90}{\scriptsize ft-qe-ctl-wmt-all} &
\rotatebox{90}{\scriptsize ft-qe-ctl-wmt-da} &
\rotatebox{90}{\scriptsize tr-qe-ctl} &
\rotatebox{90}{\scriptsize tr-qe-ctl-wmt-all} &
\rotatebox{90}{\scriptsize tr-qe-ctl-wmt-da} \\
\midrule
eng & ban & 481 & 0.56 & 0.61 & \textbf{0.62} & 0.60 & 0.54 & 0.56 & 0.57 \\
eng & guj & 401 & 0.36 & \textbf{0.38} & 0.37 & \textbf{0.38} & 0.37 & 0.34 & \textbf{0.38} \\
eng & hin & 571 & 0.57 & \textbf{0.60} & \textbf{0.60} & \textbf{0.60} & 0.56 & 0.59 & 0.58 \\
eng & kan & 559 & 0.53 & 0.59 & \textbf{0.60} & \textbf{0.60} & 0.54 & 0.55 & 0.56 \\
eng & kas & 491 & 0.23 & 0.48 & 0.50 & \textbf{0.51} & 0.43 & 0.49 & 0.47 \\
eng & mar & 598 & 0.57 & \textbf{0.63} & 0.61 & 0.62 & 0.59 & 0.60 & 0.59 \\
eng & odi & 103 & 0.41 & \textbf{0.47} & 0.46 & \textbf{0.47} & 0.45 & \textbf{0.47} & \textbf{0.47} \\
eng & pan & 349 & 0.48 & 0.53 & \textbf{0.54} & \textbf{0.54} & 0.50 & 0.52 & 0.50 \\
eng & tam & 538 & 0.58 & \textbf{0.62} & 0.61 & \textbf{0.62} & 0.56 & 0.56 & 0.56 \\
eng & tel & 615 & 0.54 & 0.59 & \textbf{0.60} & 0.59 & 0.55 & 0.56 & 0.58 \\
eng & urd & 525 & 0.39 & \textbf{0.41} & \textbf{0.41} & \textbf{0.41} & 0.38 & 0.39 & 0.39 \\
hin & ban & 274 & 0.38 & \textbf{0.50} & 0.49 & \textbf{0.50} & 0.44 & 0.49 & 0.47 \\
hin & doi & 451 & 0.02 & 0.61 & \textbf{0.62} & 0.61 & 0.49 & 0.53 & 0.55 \\
hin & guj & 379 & 0.30 & 0.32 & 0.32 & \textbf{0.33} & 0.31 & 0.30 & 0.32 \\
hin & kan & 525 & 0.43 & \textbf{0.54} & 0.52 & \textbf{0.54} & 0.53 & 0.53 & \textbf{0.54} \\
hin & mar & 597 & 0.55 & \textbf{0.66} & \textbf{0.66} & \textbf{0.66} & 0.59 & 0.61 & 0.61 \\
hin & odi & 456 & 0.25 & 0.32 & \textbf{0.33} & \textbf{0.33} & 0.30 & \textbf{0.33} & \textbf{0.33} \\
hin & pan & 513 & 0.44 & 0.55 & \textbf{0.57} & \textbf{0.57} & 0.50 & 0.51 & 0.53 \\
hin & snd & 448 & 0.45 & 0.55 & 0.56 & \textbf{0.57} & 0.51 & 0.54 & 0.53 \\
hin & tel & 403 & 0.42 & 0.48 & 0.48 & \textbf{0.49} & 0.45 & 0.46 & 0.45 \\
hin & urd & 610 & 0.43 & 0.49 & 0.52 & \textbf{0.53} & 0.47 & 0.49 & 0.50 \\
\midrule
all$^{*}$ & all & 9887 & 0.42 & 0.52 & 0.52 & 0.53 & 0.48 & 0.50 & 0.50 \\
all$^{\dag}$ & all & 9887 & 0.43 & 0.53 & 0.53 & \textbf{0.54} & 0.49 & 0.51 & 0.51 \\
\bottomrule
\end{tabular*}
\end{minipage}
\end{table}

We can see in Table \ref{tbl:comtail-qe} that as with the DA models, it is the fine-tuned model, fine-tuned with combined COMTAIL and WMT-DA dataset that comes out on top. This model with a correlation value of \textbf{0.54} compared to the best COMET model at $0.43$ achieves $25\%$ improvement. 17/21 language pairs register highest correlation values for this particular model. Interestingly, the difference between trained and fine-tuned models is greater amongst the QE models than the DA models. This points to the mixed quality of \textit{gold} reference data. 

Also while fine-tuning, the COMTAIL and WMT-DA dataset combined yields the best model, although the difference is not significant across the 21 langauges on a paired t-test (paired t-test, p = 0.023, p < 0.01). This shows that the base model used for fine-tuning had already learned from the WMT-DA datatsets, thus fine-tuning with that data included does not greatly benefit the model.

Comparing the best DA model from Table \ref{tbl:comtail-da} and the best QE model from Table \ref{tbl:comtail-qe}, we see the QE model (mean $\tau$ = 0.54) significantly outperform the DA model (mean $\tau$ = 0.53) across 21 language pairs (paired t-test, p = 0.009, p < 0.01). This is not surprising and is in line with similar previous observations \citep{rei2021references}.

\subsection{LLAMA Models}
We now turn our attention to the LLM-based models. Recall that owing to compute constraints, we only fine-tune the Llama models with the COMTAIL dataset. However, we use the scores first in their raw form on a 0--100 scale, and then in their normalized form with min-max normalization on a 0--1 scale.

Table \ref{tbl:llama-da-qe} shows the results from both the methods. We report both the DA and QE models in the same table. The Llama-based DA model fine-tuned with raw scores from the COMTAIL dataset is the best performing model amongst the four models trained in this setup. Surprisingly, the QE model trained with normalized scores has very low correlation values compared to the QE model trained on raw scores. Using raw scores does appear to help train a better model instead of normalized, min-maxed scores. These results again establish that LLM-based models are increasingly at par with SOTA COMET-based models on the evaluation scoring task \citep{mujadiaLargeLanguageModel2024}. 

\begin{table}[h]
\caption{Llama DA and QE Models: Kendall’s Tau ($\tau$) correlations obtained for each language pair on the COMTAIL test set. ${*}$ marked row lists the simple average of the correlations across all language pairs that make up the test set. ${\dag}$ marked row lists the weighted average of the correlations as per each language pair's contribution to the total test set.}
\label{tbl:llama-da-qe}
\small
\centering
\begin{tabular}{p{0.05\linewidth}p{0.05\linewidth}p{0.05\linewidth}p{0.1\linewidth}p{0.1\linewidth}p{0.1\linewidth}p{0.1\linewidth}}
\toprule
\scriptsize srclng & \scriptsize tgtlng & \scriptsize count  & \scriptsize da-raw-llama3.1 & \scriptsize da-mx-llama3.1 & \scriptsize qe-raw-llama3.1 & \scriptsize qe-mx-llama3.1 \\
\midrule
eng & ban & 481  & \textbf{0.61} & 0.57 & 0.59 & 0.51 \\
eng & guj & 401  & 0.39 & 0.38 & \textbf{0.40} & 0.38 \\
eng & hin & 571  & \textbf{0.61} & 0.60 & 0.60 & 0.57 \\
eng & kan & 559  & \textbf{0.60} & 0.58 & 0.59 & 0.56 \\
eng & kas & 491  & \textbf{0.50} & 0.46 & 0.45 & 0.09 \\
eng & mar & 598  & \textbf{0.61} & \textbf{0.61} & 0.60 & 0.56 \\
eng & odi & 103  & \textbf{0.49} & \textbf{0.49} & 0.45 & 0.47 \\
eng & pan & 349  & \textbf{0.55} & 0.52 & 0.52 & 0.48 \\
eng & tam & 538  & 0.60 & \textbf{0.61} & 0.59 & 0.57 \\
eng & tel & 615  & \textbf{0.61} & 0.58 & 0.54 & 0.55 \\
eng & urd & 525  & \textbf{0.42} & \textbf{0.42} & 0.40 & 0.38 \\
hin & ban & 274  & 0.50 & 0.47 & \textbf{0.54} & 0.48 \\
hin & doi & 451  & \textbf{0.58} & 0.50 & 0.52 & -0.05 \\
hin & guj & 379  & 0.27 & 0.28 & 0.28 & 0.25 \\
hin & kan & 525  & \textbf{0.60} & 0.56 & 0.54 & 0.53 \\
hin & mar & 597  & 0.60 & 0.59 & \textbf{0.62} & 0.57 \\
hin & odi & 456  & 0.32 & 0.31 & \textbf{0.34} & 0.32 \\
hin & pan & 513  & \textbf{0.59} & 0.55 & 0.56 & 0.51 \\
hin & snd & 448  & 0.52 & 0.48 & \textbf{0.54} & 0.38 \\
hin & tel & 403  & 0.47 & 0.47 & \textbf{0.48} & \textbf{0.48} \\
hin & urd & 610  & \textbf{0.53} & \textbf{0.53} & 0.52 & 0.51 \\
\midrule
all$^{*}$ & all & 9887  & 0.52 & 0.50 & 0.51 & 0.43 \\
all$^{\dag}$ & all & 9887  & \textbf{0.53} & 0.51 & 0.52 & 0.44 \\
\bottomrule
\end{tabular}
\end{table}

\subsection{Ablations}

\subsubsection{Volume}

We now look at the results from our volume ablation experiment. Table \ref{tbl:abl-volume-result} shows that starting with an average correlation value of $0.44$ as the baseline for the public COMET-DA model, as we fine-tune with the COMTAIL dataset in proportions of 10\%, 25\%, 50\%, 75\%, and finally all of the data at 100\%, we see a gradual improvement in correlations values until 75\% of the data, after which the gains flatten out at \textbf{0.51}. This establishes that the amount of ratings data created under the exercise was close to the minimum needed to build a stable metric model for these language pairs.

For comparison the COMTAIL \textit{trained} model's scores are also shown for the entire training dataset.

\begin{table}[h]
\caption{Volume Ablation Models. Kendall’s Tau ($\tau$) correlations obtained for each language pair on the COMTAIL test set. ${*}$ denotes row with simple average. ${\dag}$ denotes row with weighted average.}
\label{tbl:abl-volume-result}
\footnotesize
\centering
\begin{tabular*}{\textwidth}{@{\extracolsep{\fill}}p{0.03\textwidth}p{0.03\textwidth}p{0.03\textwidth}p{0.03\textwidth}p{0.05\textwidth}p{0.03\textwidth}p{0.03\textwidth}p{0.03\textwidth}p{0.03\textwidth}p{0.03\textwidth}p{0.03\textwidth}p{0.03\textwidth}p{0.03\textwidth}p{0.03\textwidth}p{0.03\textwidth}p{0.03\textwidth}}
\toprule
\rotatebox{90}{\scriptsize src-lng} &
\rotatebox{90}{\scriptsize tgt-lng} &
\rotatebox{90}{\scriptsize count} &
\rotatebox{90}{\scriptsize bleu} &
\rotatebox{90}{\scriptsize ter} &
\rotatebox{90}{\scriptsize chrf2} &
\rotatebox{90}{\scriptsize comet-wmt22-da} &
\rotatebox{90}{\scriptsize comet-xl} &
\rotatebox{90}{\scriptsize da-ai4b} &
\rotatebox{90}{\scriptsize ft-da-ctl-10} &
\rotatebox{90}{\scriptsize ft-da-ctl-25} &
\rotatebox{90}{\scriptsize ft-da-ctl-50} &
\rotatebox{90}{\scriptsize ft-da-ctl-75} &
\rotatebox{90}{\scriptsize ft-da-ctl-100} &
\rotatebox{90}{\scriptsize tr-comtail-ctl} \\
\midrule
eng & ban & 481 & 0.28 & -0.31 & 0.26 & 0.54 & 0.53 & 0.46 & 0.55 & 0.56 & 0.57 & \textbf{0.58} & \textbf{0.58} & \textbf{0.58} \\
eng & guj & 401 & 0.19 & -0.23 & 0.17 & 0.33 & 0.34 & 0.29 & 0.34 & 0.34 & 0.34 & 0.35 & \textbf{0.36} & \textbf{0.36} \\
eng & hin & 571 & 0.20 & -0.23 & 0.15 & 0.47 & 0.53 & 0.48 & 0.49 & 0.51 & 0.52 & 0.55 & 0.55 & \textbf{0.56} \\
eng & kan & 559 & 0.34 & -0.38 & 0.31 & 0.52 & 0.51 & 0.49 & 0.54 & 0.54 & 0.56 & \textbf{0.59} & \textbf{0.59} & \textbf{0.59} \\
eng & kas & 491 & 0.36 & -0.31 & 0.37 & 0.39 & 0.06 & 0.29 & 0.43 & 0.44 & 0.45 & \textbf{0.49} & \textbf{0.49} & \textbf{0.49} \\
eng & mar & 598 & 0.37 & -0.41 & 0.34 & 0.56 & 0.52 & 0.52 & 0.58 & 0.59 & 0.59 & \textbf{0.62} & \textbf{0.62} & 0.61 \\
eng & odi & 103 & 0.29 & -0.35 & 0.30 & 0.43 & 0.43 & 0.44 & 0.44 & 0.45 & 0.46 & 0.47 & \textbf{0.48} & \textbf{0.48} \\
eng & pan & 349 & 0.33 & -0.34 & 0.33 & 0.44 & 0.48 & 0.41 & 0.46 & 0.48 & 0.48 & 0.50 & 0.50 & \textbf{0.51} \\
eng & tam & 538 & 0.32 & -0.31 & 0.33 & 0.51 & 0.48 & 0.46 & 0.52 & 0.53 & 0.54 & 0.57 & 0.57 & \textbf{0.58} \\
eng & tel & 615 & 0.35 & -0.37 & 0.34 & 0.52 & 0.52 & 0.51 & 0.53 & 0.54 & 0.55 & \textbf{0.58} & \textbf{0.58} & \textbf{0.58} \\
eng & urd & 525 & 0.32 & -0.31 & 0.31 & 0.40 & 0.39 & 0.38 & 0.39 & 0.40 & 0.40 & 0.40 & 0.41 & \textbf{0.42} \\
hin & ban & 274 & 0.08 & -0.09 & 0.07 & 0.32 & 0.32 & 0.29 & 0.35 & 0.38 & 0.39 & \textbf{0.44} & 0.42 & \textbf{0.44} \\
hin & doi & 451 & 0.54 & -0.52 & 0.52 & 0.50 & 0.10 & 0.42 & 0.58 & 0.60 & 0.64 & 0.64 & \textbf{0.66} & 0.64 \\
hin & guj & 379 & 0.24 & -0.25 & 0.19 & 0.30 & 0.29 & 0.25 & \textbf{0.31} & \textbf{0.31} & 0.30 & 0.30 & 0.30 & 0.30 \\
hin & kan & 525 & 0.31 & -0.35 & 0.28 & 0.49 & 0.46 & 0.47 & 0.51 & 0.52 & 0.52 & 0.54 & \textbf{0.55} & \textbf{0.55} \\
hin & mar & 597 & 0.17 & -0.18 & 0.16 & 0.43 & 0.45 & 0.43 & 0.50 & 0.52 & 0.52 & \textbf{0.59} & 0.58 & \textbf{0.59} \\
hin & odi & 456 & 0.21 & -0.22 & 0.20 & 0.27 & 0.26 & 0.26 & 0.28 & 0.29 & 0.29 & 0.31 & 0.31 & \textbf{0.32} \\
hin & pan & 513 & 0.35 & -0.36 & 0.32 & 0.51 & 0.50 & 0.46 & 0.52 & 0.52 & 0.54 & 0.53 & \textbf{0.55} & \textbf{0.55} \\
hin & snd & 448 & 0.27 & -0.29 & 0.27 & 0.39 & 0.41 & 0.39 & 0.49 & 0.51 & 0.53 & 0.54 & \textbf{0.55} & \textbf{0.55} \\
hin & tel & 403 & 0.29 & -0.31 & 0.28 & 0.41 & 0.38 & 0.40 & 0.41 & 0.42 & 0.42 & 0.45 & 0.45 & \textbf{0.46} \\
hin & urd & 610 & 0.17 & -0.17 & 0.15 & 0.35 & 0.45 & 0.37 & 0.38 & 0.41 & 0.43 & \textbf{0.48} & 0.47 & 0.47 \\
\midrule
all$^{*}$ & all & 9887 & 0.28 & -0.30 & 0.27 & 0.43 & 0.40 & 0.40 & 0.46 & 0.47 & 0.48 & 0.50 & 0.50 & 0.51 \\
all$^{\dag}$ & all & 9887 & 0.29 & -0.30 & 0.27 & 0.44 & 0.41 & 0.41 & 0.47 & 0.48 & 0.49 & 0.51 & 0.51 & \textbf{0.52} \\
\bottomrule
\end{tabular*}
\end{table}

\subsubsection{Domain}

Ablating for the three constituent domains that make up the entire COMTAIL test set, throws up some interesting results. Recall that we had only ablated for the governance and health domains. As can be seen in Table \ref{tbl:abl-domain-result}, models trained in one domain, when used for inferencing a test set of another domain register a significant drop in correlation values: to $0.46$ from a high of $0.49$ for the governance domain models and to $0.47$ from a high of $0.52$ amongst the health domain models. Both training as well as fine-tuning strategies yield the same results.

\begin{table}[h]
\caption{Domain DA Ablation Models. Kendall’s Tau ($\tau$) correlations obtained for each domain's test set compared with entire COMTAIL test set. }
\label{tbl:abl-domain-result}
\small
\centering
\begin{tabular*}{\textwidth}{@{\extracolsep{\fill}}ccccccccccc} 
\toprule
\rotatebox{90}{\scriptsize test-set} &
\rotatebox{90}{\scriptsize count} &
\rotatebox{90}{\scriptsize comet-wmt22-da} &
\rotatebox{90}{\scriptsize comet-xl} &
\rotatebox{90}{\scriptsize da-ai4b} &
\rotatebox{90}{\scriptsize ft-comtail-iiith} &
\rotatebox{90}{\scriptsize tr-comtail-iiith} &
\rotatebox{90}{\scriptsize ft-da-abl-gov} &
\rotatebox{90}{\scriptsize ft-da-abl-hlt} &
\rotatebox{90}{\scriptsize tr-da-abl-gov} &
\rotatebox{90}{\scriptsize tr-da-abl-hlt} \\
\midrule
general & 4010 & 0.47 & 0.43 & 0.42 & \textbf{0.51} & \textbf{0.51} & 0.47 & 0.48 & 0.47 & 0.48 \\
\rowcolor{gray!20} governance & 3247 & 0.41 & 0.40 & 0.38 & \textbf{0.49} & \textbf{0.49} & \textbf{0.49} & 0.46 & \textbf{0.49} & 0.46 \\
\rowcolor{gray!20} health & 2630 & 0.43 & 0.39 & 0.42 & \textbf{0.53} & \textbf{0.53} & 0.47 & 0.52 & 0.47 & 0.52 \\
comtail & 9887 & 0.44 & 0.41 & 0.41 & 0.51 & \textbf{0.52} & 0.48 & 0.49 & 0.48 & 0.49 \\
\bottomrule
\end{tabular*}
\end{table}

\subsubsection{Quality}

Table \ref{tbl:abl-quality-result} shows the results for ablation on quality subsets. It is noteworthy that the \textit{good} quality test subset registers much lower correlation values across the board than the \textit{neutral} and \textit{bad} subsets. This reinforces the point that it is harder to discriminate between higher quality translations than translations of lower quality. Do models fine-tuned or trained on only \textit{bad} or passable quality data struggle more than the \textit{good} data quality models? It does not appear so. This is an important point to note for any future data creation tasks. Should data primarily be made up of high quality translations having been rated which allows the model to learn nuanced discrimination of quality? Our results show that all quality gradations contribute to the robustness of the model. As to what constitutes the optimal volume from each category remains to be determined.

QE models built and tested with the same data setup as shown in Table \ref{tbl:abl-quality-qe-result} exhibit a similar trend.

\begin{table}[h]
\caption{Quality DA Ablation Models. Kendall’s Tau ($\tau$) correlations obtained for each quality category's test set compared with entire COMTAIL test set.}
\label{tbl:abl-quality-result}
\footnotesize
\centering
\begin{tabular*}{\textwidth}{@{\extracolsep{\fill}}ccccccccccccc} 
\toprule
\rotatebox{90}{\scriptsize test-set} &
\rotatebox{90}{\scriptsize count} &
\rotatebox{90}{\scriptsize ft-da-bad} &
\rotatebox{90}{\scriptsize ft-da-good} &
\rotatebox{90}{\scriptsize ft-da-neutral} &
\rotatebox{90}{\scriptsize tr-da-bad} &
\rotatebox{90}{\scriptsize tr-da-good} &
\rotatebox{90}{\scriptsize tr-da-neutral} &
\rotatebox{90}{\scriptsize comet-wmt22-da} &
\rotatebox{90}{\scriptsize comet-xl} &
\rotatebox{90}{\scriptsize da-ai4b} &
\rotatebox{90}{\scriptsize ft-comtail} &
\rotatebox{90}{\scriptsize tr-comtail} \\
\midrule
good & 4107 & 0.23 & 0.28 & 0.23 & 0.19 & 0.27 & 0.19 & 0.19 & 0.24 & 0.20 & 0.28 & \textbf{0.29} \\
bad & 2281 & \textbf{0.51} & 0.46 & 0.44 & 0.49 & 0.41 & 0.42 & 0.43 & 0.33 & 0.36 & \textbf{0.51} & \textbf{0.51} \\
neutral & 3499 & 0.43 & 0.43 & 0.49 & 0.41 & 0.39 & 0.47 & 0.44 & 0.38 & 0.39 & \textbf{0.50} & 0.49 \\
comtail & 9887 & 0.47 & 0.46 & 0.47 & 0.45 & 0.44 & 0.45 & 0.44 & 0.41 & 0.41 & 0.51 & \textbf{0.52} \\
\bottomrule
\end{tabular*}

\end{table}

\subsubsection{Language}

We now discuss the various language ablation results. We keep our discussion to the DA models only. Results from corresponding QE models are listed in Appendix \ref{app:results}. Any noteworthy deviations between the two for different ablations are highlighted.

Table \ref{tbl:abl-language-xil-result} compares the results of the two source directions when subset and then trained and inferred separately. We find that models exclusively trained or fine-tuned on \textit{eng-il} direction training subset can reach a correlation value of $0.44$ on an \textit{hin-il} test set against a possible best of $0.50$ when the entire COMTAIL set is used.

Similarly, for the hindi source, models exclusively trained or fine-tuned on \textit{hin-il} direction training subset can reach a correlation value of $0.49$ on an \textit{eng-il} test set against a best of $0.54$, possible in this case with the \textit{eng-il} subset fine-tuned model. The drop when cross-inferencing test sets exclusive to each source is similar in terms of absolute correlation values. This to us, implies that it is the target language data that are key to improving model correlations than the source language. And if the base model in case of fine-tuning already has a substantial amount of english source data, then for the Indian language scenario it might be better to create ratings data exclusively between Indian languages.

\begin{table}[h]
\caption{Language-xil DA Ablation Models. Kendall’s Tau ($\tau$) correlations obtained for each source paired test set compared with entire COMTAIL test set.}
\label{tbl:abl-language-xil-result}
\small
\centering
\begin{tabular*}{\textwidth}{@{\extracolsep{\fill}}ccccccccccc} 
\toprule
{\scriptsize \rotatebox{90}{test-set}} &
{\scriptsize \rotatebox{90}{count}} &
{\scriptsize \rotatebox{90}{ft-da-enil}} &
{\scriptsize \rotatebox{90}{ft-da-hiil}} &
{\scriptsize \rotatebox{90}{tr-da-enil}} &
{\scriptsize \rotatebox{90}{tr-da-hiil}} &
{\scriptsize \rotatebox{90}{comet-wmt22-da}} &
{\scriptsize \rotatebox{90}{comet-xl}} &
{\scriptsize \rotatebox{90}{da-ai4b}} &
{\scriptsize \rotatebox{90}{ft-comtail}} &
{\scriptsize \rotatebox{90}{tr-comtail}} \\
\midrule
eng-il & 5231 & \textbf{0.54} & 0.49 & 0.51 & 0.49 & 0.47 & 0.44 & 0.44 & 0.53 & 0.53 \\
hin-il & 4656 & 0.44 & 0.47 & 0.42 & 0.47 & 0.40 & 0.37 & 0.38 & 0.49 & \textbf{0.50} \\
comtail & 9887 & 0.49 & 0.48 & 0.47 & 0.48 & 0.44 & 0.41 & 0.41 & 0.51 & \textbf{0.52} \\
\bottomrule
\end{tabular*}
\end{table}

\subsubsection{Language-family}

How do the models trained exclusively on the two language families \textit{Indo-Aryan} and \textit{Dravidian} fare when put through ablated test data? 

As can be seen in Table \ref{tbl:abl-language-family-result}, for the \textit{dravidian} test set, the best possible correlation value is $0.56$, against which, the model trained only with \textit{indo-aryan} data achieves a value of $0.53$. Similarly, for the \textit{indo-aryan} test set the best possible correlation value is $0.50$ against which the model trained only with \textit{dravidian} data achieves a value of $0.44$, a larger quantum of difference. Can this be taken as evidence of greater diffusion from Indo-aryan into Dravidian than the other way round? But also noticeable is the volume difference in the two training subsets with Indo-aryan training subset being 3 times larger than the Dravidian subset. However, the results clearly imply that differences between the two families are not that stark, and that each family  is aided by the other in a multilingual learning setup.

\begin{table}[h]
\caption{Language-family DA Ablation Models. Kendall’s Tau ($\tau$) correlations obtained for each language family's test set compared with entire COMTAIL test set.}
\label{tbl:abl-language-family-result}
\small
\centering
\begin{tabular*}{\textwidth}{@{\extracolsep{\fill}}ccccccccccc} 
\toprule
{\scriptsize \rotatebox{90}{test-set}} &
{\scriptsize \rotatebox{90}{count}} &
{\scriptsize \rotatebox{90}{ft-da-ary}} &
{\scriptsize \rotatebox{90}{ft-da-drv}} &
{\scriptsize \rotatebox{90}{tr-da-ary}} &
{\scriptsize \rotatebox{90}{tr-da-drv}} &
{\scriptsize \rotatebox{90}{comet-wmt22-da}} &
{\scriptsize \rotatebox{90}{comet-xl}} &
{\scriptsize \rotatebox{90}{da-ai4b}} &
{\scriptsize \rotatebox{90}{ft-comtail}} &
{\scriptsize \rotatebox{90}{tr-comtail}} \\
\midrule
dravidian & 2640 & 0.53 & \textbf{0.56} & 0.51 & 0.52 & 0.49 & 0.48 & 0.47 & 0.55 & \textbf{0.56} \\
indo-aryan & 7247 & 0.49 & 0.44 & 0.48 & 0.41 & 0.42 & 0.39 & 0.39 & \textbf{0.50} & \textbf{0.50} \\
comtail & 9887 & 0.50 & 0.47 & 0.49 & 0.44 & 0.44 & 0.41 & 0.41 & 0.51 & \textbf{0.52} \\
\bottomrule
\end{tabular*}

\end{table}

\subsubsection{Language-subfamily}

Table \ref{tbl:abl-language-subfamily-result} shows the correlation values on the subfamily test sets for models built on ablated data. In each row, the values in \textbf{bold} are the best correlation values obtained on that subfamilies test data, the \underline{underlined} values in the same row represent the second best models, while the \textit{italicized} values show the third best. This is done to be able to discover as to which subfamily offers the best possible model in the absence of a model trained on a particular subfamily's data. We only note the subfamily models and ignore the public and comtail model trained on the full dataset, although the table represents them for completeness.  

Analysing the results in this manner, we note that for the two dravidian subfamilies, the best possible subfamily models come from indo-aryan subfamilies, namely: ia-central and ia-east, in that order for dv-south (tamil and kannada); and dv-south, ia-central, ia-east, and ia-north-west for dv-south-central (telugu).

Amongst the indo-aryan subfamilies, ia-south (marathi) appears to do well with dv-south (tamil, kannada); ia-dardic (kashmiri), always difficult to classify within the indian lingusitic area, does well on ia-north (dogri) model; also surprising that a script overlap with urdu or sindhi (both Perso-Arabic based) does not seem to be of much help. Many of these results speak to diffusion across genetic families and testify to geographical contiguity playing an important role. 

These are important factors to consider at the time of data collection or experiment design when working with an array of related languages or languages found within a linguistic area. We acknowledge a similar observations also made by \cite{singhHowGoodZeroShot2024} in a low resource Indian language setting.

\begin{table}[h]
\caption{Language-subfamily DA Ablation Models. Kendall’s Tau ($\tau$) correlations obtained for each subfamily's test set compared with entire COMTAIL test set. In each row, the values in \textbf{bold} are the best correlation values obtained on that subfamily's test data, the \underline{underlined} values in the same row represent the second best models, while the \textit{italicized} values show the third best}.
\label{tbl:abl-language-subfamily-result}
\footnotesize
\centering
\begin{tabular*}{\textwidth}{p{0.08\linewidth}p{0.03\linewidth}p{0.03\linewidth}p{0.03\linewidth}p{0.03\linewidth}p{0.03\linewidth}p{0.03\linewidth}p{0.03\linewidth}p{0.03\linewidth}p{0.03\linewidth}p{0.03\linewidth}p{0.03\linewidth}p{0.04\linewidth}p{0.03\linewidth}p{0.03\linewidth}}
\toprule
test-set & count & ft-dv-south & ft-dv-south-central & ft-ia-central & ft-ia-dardic & ft-ia-east & ft-ia-north & ft-ia-north-west & ft-ia-south & ft-ia-west & comet-wmt22-da & comet-xl & da-ai4b & ft-comtail \\
\midrule
dv-south & 1622 & \underline
{0.55} & 0.51 & \underline{0.55} & 0.50 & \textit{0.54} & 0.45 & 0.54 & 0.49 & 0.51 & 0.51 & 0.48 & 0.47 & \textbf{0.57} \\
dv-south-central & 1018 & \underline{0.50} & \underline{0.50} & \underline{0.50} & 0.48 & \underline{0.50} & 0.44 & \textit{0.49} & 0.46 & 0.46 & 0.48 & 0.47 & 0.46 & \textbf{0.53} \\
ia-central & 1706 & \textit{0.47} & 0.40 & \textbf{0.49} & 0.41 & 0.42 & 0.36 & 0.44 & 0.43 & 0.42 & 0.41 & 0.46 & 0.41 & \underline{0.48} \\
ia-dardic & 491 & 0.35 & 0.25 & 0.27 & \underline{0.42} & 0.38 & \textit{0.40} & 0.39 & 0.35 & 0.36 & 0.39 & 0.06 & 0.29 & \textbf{0.49} \\
ia-east & 1314 & \underline{0.42} & 0.38 & \textit{0.40} & 0.38 & \textbf{0.45} & 0.35 & \textit{0.40} & 0.38 & \textit{0.40} & 0.39 & 0.38 & 0.35 & \textbf{0.45} \\
ia-north & 451 & 0.31 & 0.26 & 0.20 & \textit{0.52} & 0.45 & \textbf{0.66} & \underline{0.53} & 0.37 & 0.44 & 0.50 & 0.10 & 0.42 & \textbf{0.66} \\
ia-north-west & 1310 & \textit{0.47} & 0.46 & \textit{0.47} & 0.43 & 0.45 & 0.45 & \underline{0.53} & \textit{0.47} & 0.45 & 0.45 & 0.46 & 0.42 & \textbf{0.54} \\
ia-south & 1195 & \textit{0.57} & 0.51 & 0.56 & 0.48 & 0.53 & 0.49 & \textit{0.57} & \underline{0.59} & 0.53 & 0.49 & 0.49 & 0.47 & \textbf{0.60} \\
ia-west & 780 & \underline{0.34} & \textit{0.33} & \textit{0.33} & 0.30 & 0.32 & 0.31 & \textit{0.33} & \underline{0.34} & \textbf{0.35} & 0.31 & 0.32 & 0.27 & \textit{0.33} \\
comtail & 9887 & 0.47 & 0.43 & 0.46 & 0.44 & 0.46 & 0.42 & 0.48 & 0.45 & 0.44 & 0.44 & 0.41 & 0.41 & \textbf{0.52} \\
\bottomrule
\end{tabular*}
\end{table}

\subsubsection{Language-Hindi-Urdu}

The final table under our ablation experiments delves into the Hindi-Urdu debate. If indeed it is one language and two scripts then either model should presumably work for the other. 

The results from Table \ref{tbl:abl-language-hin-urd-result} do not appear to support this claim. While the eng-hin test set can be adjudged with a high correlation value ($0.58$) by the model fine-tuned in this direction and a much lower value ($0.48$) by the eng-urd model, the eng-urd test set can be inferred almost equally well by both the eng-hin and eng-urd fine-tuned models ($0.39$ and $0.40$ respectively). Why is hindi difficult for the urdu model and not the other way round? Is it the case that urdu has diffused more into hindi, primarily in terms of vocabulary? Or, is it a matter of the ratings data quality in each of these languages? Comparing against the hin-urd test set, we find a similar trend.

We note that previous efforts in this direction, although not on a similar task, had taken a script normalization approach to arrive at a common dataset and shown mutual gains \citep{bhat-etal-2016-house, visweswariah2010urdu}. However, in our case, because of the underlying multilingual pre-trained model, it has been possible to put this to test by training models for the two languages in their original scripts separately. However, we note that this particular debate needs further in-depth analysis, especially of vocabulary convergence and divergence between the two languages.

\begin{table}[h]
\caption{Language-Hindi-Urdu DA Ablation Models. Kendall’s Tau ($\tau$) correlations obtained for hindi and urdu target test sets compared with entire COMTAIL test set.}
\label{tbl:abl-language-hin-urd-result}
\small
\centering
\begin{tabular*}{\textwidth}{@{\extracolsep{\fill}}p{1.2cm}  
p{0.8cm}  
p{0.8cm}  
p{1.0cm}  
p{0.8cm}  
p{0.8cm}  
p{0.8cm}  
p{0.8cm}  
p{0.8cm}  
p{0.8cm}  
}
\toprule
test-set & count & comet-wmt22-da & comet-xl & da-ai4b & ft-comtail & ft-da-enghin & ft-da-engurd & ft-da-hinurd & tr-comtail \\
\midrule
eng-hin & 571 & 0.47 & 0.53 & 0.48 & 0.55 & \textbf{0.58} & 0.48 & 0.53 & 0.56 \\
eng-urd & 525 & 0.40 & 0.39 & 0.38 & 0.41 & 0.39 & 0.40 & 0.40 & \textbf{0.42} \\
hin-urd & 610 & 0.35 & 0.45 & 0.37 & 0.47 & 0.42 & 0.37 & \textbf{0.48} & 0.47 \\
comtail & 9887 & 0.44 & 0.41 & 0.41 & 0.51 & 0.45 & 0.45 & 0.46 & \textbf{0.52} \\
\bottomrule
\end{tabular*}

\end{table}

\subsection{Challenge Set}

We report our results on the ACES challenge set as follows: a language-pair grouped comparison and a phenomena grouped comparison.

Table \ref{tbl:abl-challenge-language-result} shows the 7 Indian language pairs evaluated using four models: a DA and a QE model from COMET and a DA and a QE model from COMTAIL. The results are mixed, the COMTAIL models score higher correlations for 4 of the 7 pairs and tie with the COMET models on 1 pair. However, we advise caution in interpreting these scores since they are affected by the variable number of items and phenomena in each pair.

\begin{table}[h]
\caption{Challenge Set by IL language-pair. Kendall’s Tau ($\tau$) correlations obtained for Indian language pairs within the ACES dataset.}
\label{tbl:abl-challenge-language-result}
\small
\centering
\begin{tabular*}{\textwidth}{@{\extracolsep{\fill}}cccccc} 
\toprule
 lang-pair & counts & cmt-da & cmt-qe & ctl-da & ctl-qe \\
\midrule
en-hi & 343 & 0.67 & 0.46 & \textbf{0.78} & -0.70 \\
en-mr & 52 & \textbf{0.88} & 0.77 & 0.77 & 0.85 \\
en-ur & 10 & 0.40 & \textbf{0.60} & \textbf{0.60} & 0.20 \\
hi-en & 367 & 0.59 & \textbf{0.77} & 0.49 & 0.75 \\
mr-en & 63 & -0.33 & -0.40 & \textbf{-0.17} & -0.40 \\
ta-en & 39 & -0.28 & -0.18 & \textbf{0.03} & -0.13 \\
ur-en & 372 & 0.50 & 0.74 & 0.30 & \textbf{0.78} \\
\bottomrule
\end{tabular*}
\end{table}

Table \ref{tbl:abl-challenge-phenomena-result} shows the phenomena grouped results. COMTAIL metrics register a higher correlation value on only 6 out 13 phenomena listed. They  struggle in discriminating particularly those phenomena which are not represented in the COMTAIL dataset, such as copy-source which is just the copying of the entire source segment as output and some of the hallucination phenomenon. Although these phenomena were likely captured by the base COMET model, we hypothesize that fine-tuning on our dataset may have led the model to forget some of the knowledge acquired from the larger WMT data.

\begin{table}[h]
\caption{Challenge Set by phenomena. Kendall’s Tau ($\tau$) correlations obtained for specific phenomena on Indian language pairs within the ACES dataset.}
\label{tbl:abl-challenge-phenomena-result}
\small
\centering
\begin{tabular*}{\textwidth}{@{\extracolsep{\fill}}p{6.0cm} c c c c c 
}
\toprule
 phenomenon & counts & cmt-da & cmt-qe & ctl-da & ctl-qe \\
 \midrule
addition & 22 & 0.00 & 0.73 & -0.09 & \textbf{0.82} \\
copy-source & 63 & \textbf{0.84} & 0.46 & 0.17 & 0.81 \\
hallucination-date-time & 28 & -0.29 & \textbf{0.50} & -0.21 & 0.29 \\
hallucination-real-data-vs-synonym & 29 & \textbf{1.00} & 0.86 & 0.86 & 0.86 \\
hallucination-unit-conversion-amount-matches-ref & 89 & -0.53 & -0.84 & \textbf{-0.30} & -0.87 \\
hallucination-unit-conversion-unit-matches-ref & 89 & -0.08 & 0.03 & \textbf{0.06} & \textbf{0.06} \\
omission & 39 & 0.64 & \textbf{0.95} & 0.54 & 0.85 \\
overly-literal-vs-synonym & 20 & 0.70 & 0.60 & 0.70 & \textbf{0.80} \\
similar-language-high & 319 & 0.69 & 0.42 & \textbf{0.82} & -0.83 \\
xnli-addition-contradiction & 147 & 0.70 & \textbf{0.99} & 0.70 & 0.97 \\
xnli-addition-neutral & 138 & 0.64 & 0.90 & 0.54 & \textbf{0.93} \\
xnli-omission-contradiction & 133 & 0.68 & \textbf{1.00} & 0.44 & 0.98 \\
xnli-omission-neutral & 131 & 0.66 & \textbf{1.00} & 0.42 & 0.97 \\
\bottomrule
\end{tabular*}

\end{table}

\section{Conclusion}
\label{sec:conclusion}

We presented the COMTAIL dataset, a machine translation evaluation dataset based on DA+SQM rating methodology, that contains 219,439 items across its training, development, and evaluation sets. The dataset contains 2 source languages paired with 13 Indian languages making up 21 unique translation directions, and to our knowledge is the largest such dataset by volume and language coverage within the Indian languages setting.
We described the methodology followed in the creation of the dataset and presented details of our rater selection, quality control, and data normalization methods. Our approach was inspired by the best practices laid out during the WMT human evaluation tasks but adapted to our conditions, for instance, introduction of the source language reading comprehension test for the rater selection task outlined in Section \ref{sec:data-raters}.

We then proceeded to build two types of COMTAIL evaluation models using this dataset by combining it with other publicly available data: COMTAIL-DA (reference-based models); and COMTAIL-QE (reference-less  models).  We found that a COMTAIL-QE model fine-tuned on top of a base COMET-QE model and using a combined dataset to significantly outperform its reference-based COMTAIL-DA counterpart as reported in Table \ref{tbl:comtail-qe} and the ensuing discussion. We also built Llama-based DA and QE models and found the Llama-based DA model to be at par with the COMTAIL models. Our foray in this direction was limited to only a few fine-tuning experiments, but this is a promising avenue given the powerful base models now available and the possibility of multi-task learning that we intend to explore \citep{alvesTowerOpenMultilingual2024}.

We then reported on a number of ablation experiments. We summarize our learning from the ablation experiments in the form of the following guidelines, hopefully useful to any future data creation or model training efforts on a related task:

\begin{enumerate}
	\item \textbf{Volume}: In our case improvements started to flatten out once we reached $75\%$ of the total data volume of our dataset. This still puts the training data required  in the order of hundreds of thousands of items for a single task setup such as ours. This is instructive for any future tasks, but the volume relationship may also be dependent on the size and nature of the underlying pre-trained model, which remains a task for future exploration.
	\item \textbf{Domain}: Trained metric models do not fare well across unseen domains. When using such models on newer domains some initial manual evaluation of the results is recommended.
	\item \textbf{Quality}: Training data must represent different translations quality gradations. Better quality human rated translations contribute more to model performance than bad quality translations. However, both are needed to build a nuanced and robust evaluation model.
	\item \textbf{Source Language}: In our experiments, the source language used in the rated translation pairs did not make as large a contribution as the target language. However, given that presently most data are English centric, we recommend a diversity of source languages to be used for any future task.
	\item \textbf{Language Families}: Indo-aryan models do well on Dravidian test data while Dravidian models do not fare as well. This is either marker of greater diffusion in one direction or indicative of a large gap in the data resources available to each family. Any future efforts should analyze this further and take it into consideration.
	\item \textbf{Language subfamilies}: We see some evidence of diffusion across families in geographically contiguous linguistic areas. For scenarios that call for zero-shot evaluation or fine-tuning with a related pair, this aspect must play  a role in the choice of model used.
	\item \textbf{Hindi-Urdu}: We noticed that models trained exclusively on each don't behave in the same manner. We put this down to one of the following: (i) Urdu possibly diffusing more into Hindi than Hindi into Urdu in our data; (ii) script differences between the two affecting inference in one language more than the other; (iii) quality differences in the two languages ratings data, Hindi being more nuanced than Urdu. The debate remains unsettled.
\end{enumerate}

Additionally, we also presented results on a publicly available challenge set. We found that apart from broad translation quality gradations and language coverage, metrics data must also contain coverage of common linguistic phenomena and their corresponding erroneous translations if robust metrics are to be built.

Time and resource constraints limited us from experimenting with Indic pre-trained models \citep{kakwani2020indicnlpsuite}, which are reported to be better suited in the Indian language scenario \citep{saibIndicMTEvalDataset2023}. Learnings from the challenge set experiments indicate that in future, we must also experiment with synthetic data augmentation techniques and introduce semantic perturbations into any datasets that we create.

Extending both the data and models to cover all 22 scheduled Indian languages across certain targeted domains remains a goal for us. Our models would also benefit from incorporating MQM and ESA based annotations in future. And, as mentioned earlier, while the COMET architecture remains competitive, greater exploration of LLM-based approaches, particularly those based on multi-task learning and explainable evaluation are some promising directions that we intend to explore.


\section*{Acknowledgements}

We thank the Ministry of Electronics and Information Technology (MeitY), Government of India for the funding and support extended under Mission Bhashini that facilitated this study.  We would like to acknowledge the National PARAM Supercomputing Facility (NPSF), C-DAC for enabling access to their Param Siddhi servers which were utilised for some of the model training experiments reported in this work. 

We thank the following colleagues from Bhashini's HIMANGY consortium for their contributions to this study and their assistance in recruiting evaluators for this task: Karunesh Arora, Aditi Mukherjee, Radhika Mamidi, Parameswari Krishnamurthy, Sunita Arora, Rajya Rama, Ashwath Rao B, Narendra VG, Annarao Kulkarni, Gurpreet Singh Lehal, Tejinder Singh, Gurpreet Singh Josan, Rakesh Chandra Balbantaray, Preeti Dubey, Adil Amin Kak, Prasenjit Majumder, Asif Ekbal, Palash Gupta, and Praveen Katamoni.

We also gratefully acknowledge the contributions of all the freelance evaluators whose work made the COMTAIL dataset possible.

\begin{appendices}

\section{Additional Details}
\label{app:details}



\subsection*{Reading Comprehension Test Samples}
\label{app:details-rc-sample}

\subsubsection*{English}

\begin{figure}[H]
  \centering
  \includegraphics[width=\textwidth]{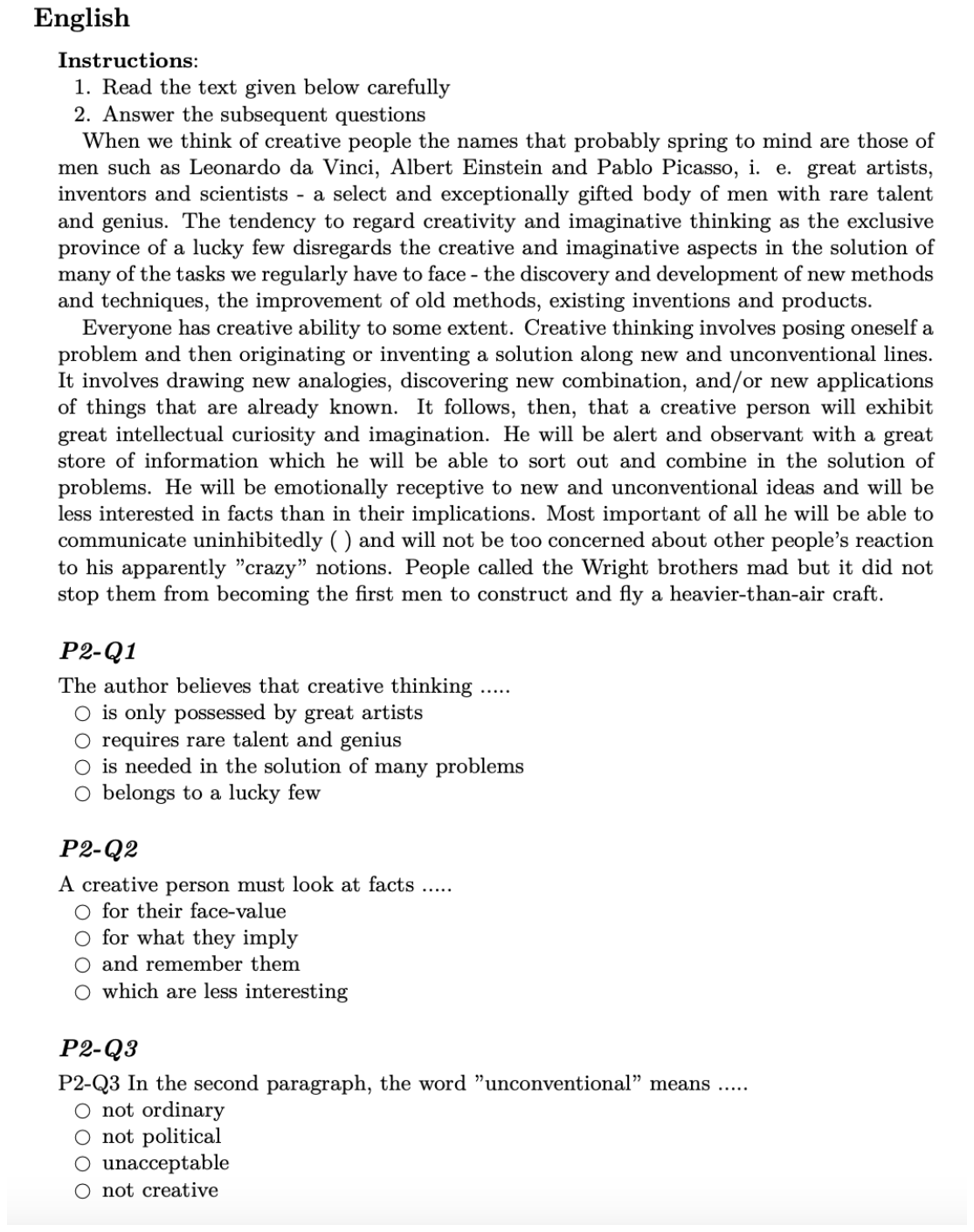}
  \caption{Sample passage from Reading Comprehension Test for English source.}
  \label{fig:rc-eng}
\end{figure}

\subsubsection*{Hindi}

\begin{figure}[H]
  \centering
  \includegraphics[width=\textwidth]{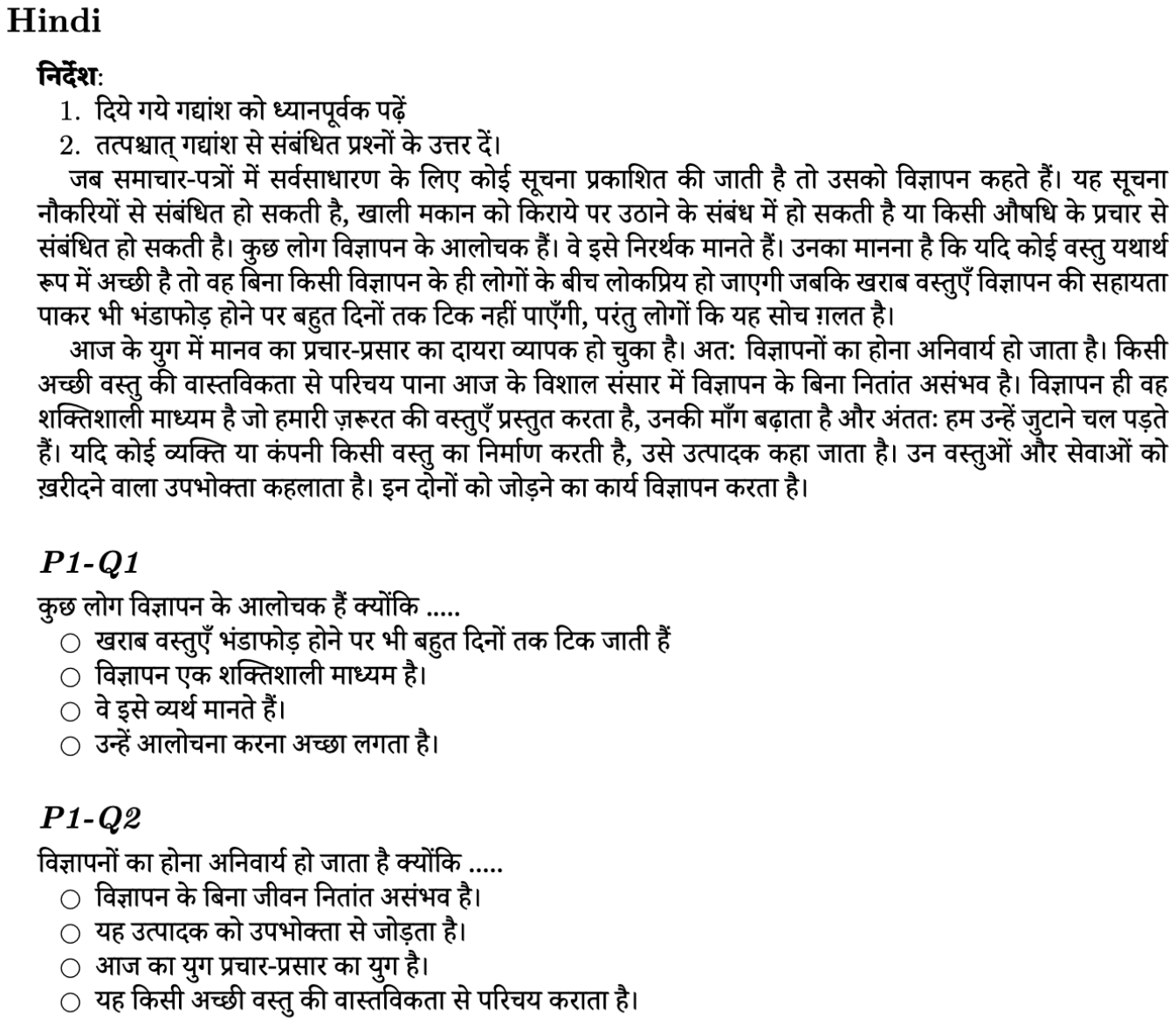}
  \caption{Sample passage from Reading Comprehension Test for Hindi source.}
  \label{fig:rc-hindi}
\end{figure}

\subsection*{Raw Data Summary}
\label{app:details-raw-summary}

\subsubsection*{}
\begin{figure}[H]
  \centering
  \includegraphics[width=0.9\textwidth]{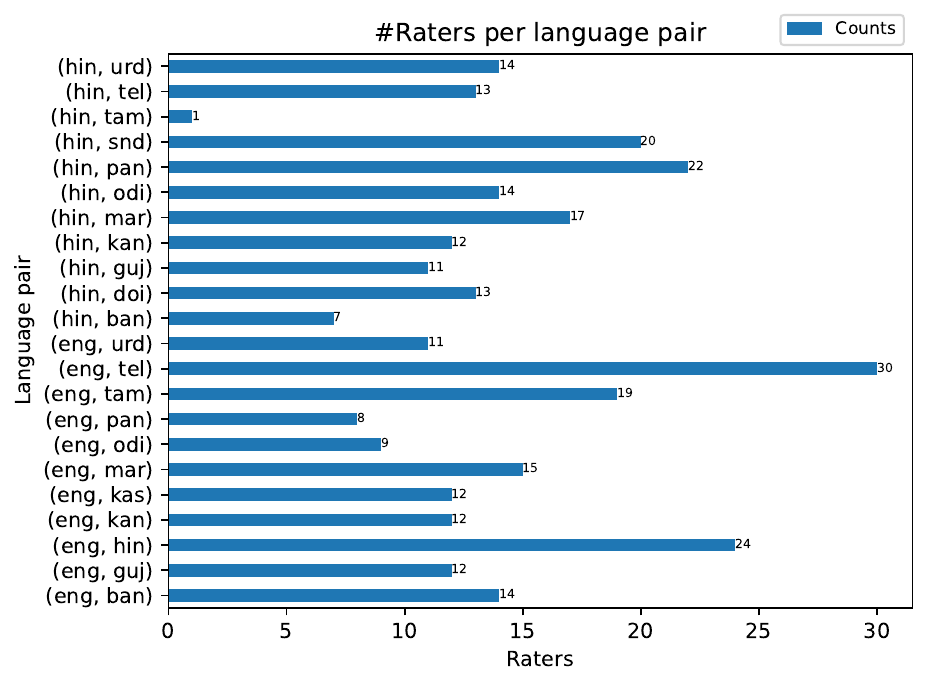}
  \caption{Number of raters per language pair.}
  \label{fig:raters_pair}
\end{figure}

\begin{figure}[H]
  \centering
  \includegraphics[width=0.9\textwidth]{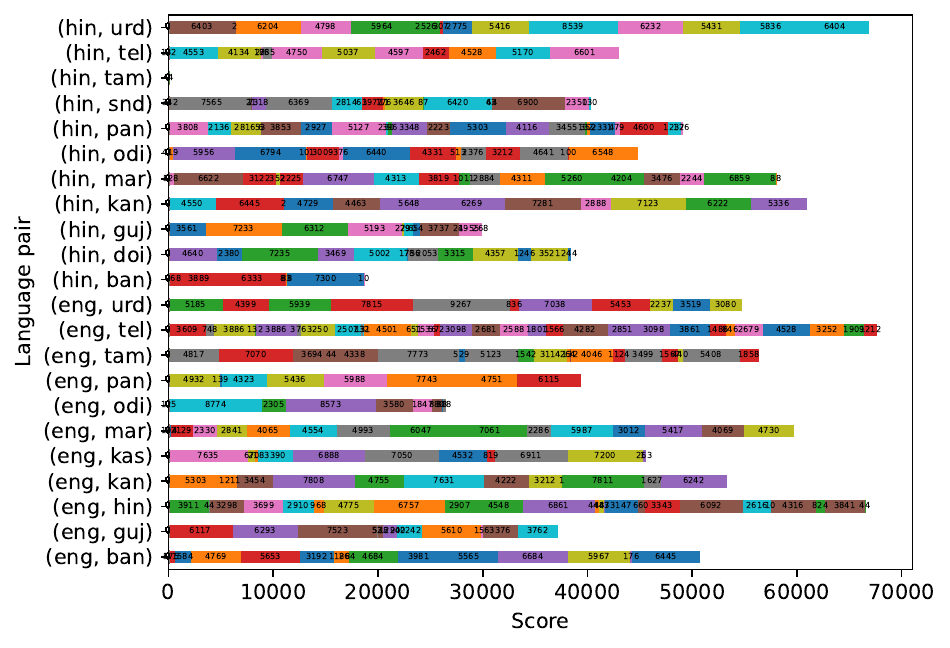}
  \caption{Rating diversity per language pair.}
  \label{fig:diversity_pair}
\end{figure}


\subsection*{Hyperparameters}
\begin{table}[h]
\centering
\caption{Hyperparameters used in training COMTAIL models. All model variants share the same set of hyper-parameters except for those explicitly marked. $^{a}$The hidden sizes of the QE models were set differently than the DA models and used $(2048, 1024)$ instead.}
\label{tab:hyperparams}
\begin{tabular*}{\textwidth}{@{\extracolsep{\fill}}ll}
\toprule
\textbf{hyperparameter} & \textbf{value} \\
\midrule
batch size       & 4 \\
dropout          & 0.1 \\
encoder learning rate    & 1.0e-06 \\
encoder model       & XLM-RoBERTa \\
hidden sizes$^{a}$     &      3072, 1024 \\
Optimizer        & AdamW \\
Epochs          & 5 \\

\hline
\end{tabular*}
\end{table}

\section{Additional Results}
\label{app:results}

\begin{table}[h]
\caption{Domain QE Ablation Models. Kendall’s Tau ($\tau$) correlations obtained for each domain's test set compared with entire COMTAIL test set. The highest correlation value for each test set row is highlighted in bold.}
\label{tbl:abl-domain-qe-result}
\small
\centering
\begin{tabular*}{\textwidth}{@{\extracolsep{\fill}}ccccccc} 
\toprule
\rotatebox{90}{\scriptsize test-set} & \rotatebox{90}{\scriptsize count} & \rotatebox{90}{\scriptsize comet-kiwi-wmt22-da} & \rotatebox{90}{\scriptsize ft-qe-abl-gov} & \rotatebox{90}{\scriptsize ft-qe-abl-hlt} & \rotatebox{90}{\scriptsize tr-qe-abl-gov} & \rotatebox{90}{\scriptsize tr-qe-abl-hlt} \\
\midrule
general & 4010 & 0.45 & \textbf{0.48} & \textbf{0.48} & 0.42 & 0.41 \\
governance & 3247 & 0.40 & \textbf{0.51} & 0.48 & 0.46 & 0.42 \\
health & 2630 & 0.44 & 0.51 & \textbf{0.53} & 0.44 & 0.48 \\
comtail & 9887 & 0.43 & \textbf{0.50} & \textbf{0.50} & 0.44 & 0.44 \\
\bottomrule
\end{tabular*}
\end{table}


\begin{table}[h]
\caption{Quality QE Ablation Models. Kendall’s Tau ($\tau$) correlations obtained for each test set compared with entire COMTAIL test set. The highest correlation value for each test set row is highlighted in bold.}
\label{tbl:abl-quality-qe-result}
\small
\centering
\begin{tabular*}{\textwidth}{@{\extracolsep{\fill}}cccccc}
\toprule
test-set & count & comet-kiwi-wmt22-da & ft-qe-bad & ft-qe-good & ft-qe-neutral \\
\midrule
comtail & 9887 & 0.43 & 0.48 & 0.48 & \textbf{0.50} \\
good & 4107 & 0.22 & 0.25 & \textbf{0.31} & 0.26 \\
bad & 2281 & 0.44 & \textbf{0.53} & 0.49 & 0.46 \\
neutral & 3499 & 0.41 & 0.46 & 0.43 & \textbf{0.50} \\
\bottomrule
\end{tabular*}
\end{table}

\begin{table}[h]
\caption{Language-xil QE Ablation Models. Kendall’s Tau ($\tau$) correlations obtained for each source direction's test set compared with entire COMTAIL test set. The highest correlation value for each test set row is highlighted in bold.}
\label{tbl:abl-language-xil-qe-result}
\footnotesize
\centering
\begin{tabular*}{\textwidth} {@{\extracolsep{\fill}}ccccccc}
\toprule
test-set & count & comet-kiwi-wmt22-da & ft-qe-enil & ft-qe-hiil & tr-qe-enil & tr-qe-hiil \\
\midrule
eng-il & 5231 & 0.49 & \textbf{0.55} & 0.50 & 0.51 & 0.44 \\
hin-il & 4656 & 0.38 & 0.38 & \textbf{0.51} & 0.36 & 0.45 \\
comtail & 9887 & 0.43 & 0.47 & \textbf{0.51} & 0.44 & 0.44 \\
\bottomrule
\end{tabular*}
\end{table}

\begin{table}[h]
\caption{Language-family QE Ablation Models. Kendall’s Tau ($\tau$) correlations obtained for each family's test set compared with entire COMTAIL test set. The highest correlation value for each test set row is highlighted in bold.}
\label{tbl:abl-language-family-qe-result}
\footnotesize
\centering
\begin{tabular*}{\textwidth} {@{\extracolsep{\fill}}ccccccc}
\toprule
test-set & count & comet-kiwi-wmt22-da & ft-qe-ary & ft-qe-drv & tr-qe-ary & tr-qe-drv \\
\midrule
dravidian & 2640 & 0.51 & 0.55 & \textbf{0.57} & 0.49 & 0.52 \\
indo-aryan & 7247 & 0.41 & \textbf{0.52} & 0.42 & 0.47 & 0.37 \\
comtail & 9887 & 0.43 & \textbf{0.53} & 0.46 & 0.47 & 0.41 \\
\bottomrule
\end{tabular*}
\end{table}


\begin{table}[h]
\caption{Language-subfamily QE Ablation Models. Kendall’s Tau ($\tau$) correlations obtained for each subfamily's test set compared with entire COMTAIL test set. The highest correlation value for each test set row is highlighted in bold.}
\label{tbl:abl-subfamily-qe-result}
\footnotesize
\centering
\begin{tabular*}{\textwidth} {@{\extracolsep{\fill}}cccccccccccc}
\toprule
\rotatebox{90}{\scriptsize test-set} &
\rotatebox{90}{\scriptsize count} &
\rotatebox{90}{\scriptsize comet-kiwi-wmt22-da} & 
\rotatebox{90}{\scriptsize ft-qe-dv-south} & 
\rotatebox{90}{\scriptsize ft-qe-dv-south-central} &
\rotatebox{90}{\scriptsize ft-qe-ia-central} &
\rotatebox{90}{\scriptsize ft-qe-ia-dardic} &
\rotatebox{90}{\scriptsize ft-qe-ia-east} & 
\rotatebox{90}{\scriptsize ft-qe-ia-north} & 
\rotatebox{90}{\scriptsize ft-qe-ia-north-west} & 
\rotatebox{90}{\scriptsize ft-qe-ia-south} &
\rotatebox{90}{\scriptsize ft-qe-ia-west} \\
\midrule
dv-south & 1622 & 0.51 & \textbf{0.58} & 0.55 & 0.54 & 0.48 & 0.55 & 0.17 & 0.55 & 0.53 & 0.52 \\
dv-south-central & 1018 & 0.49 & 0.52 & \textbf{0.55} & 0.52 & 0.44 & 0.51 & 0.23 & 0.51 & 0.49 & 0.46 \\
ia-central & 1706 & 0.46 & 0.48 & 0.48 & \textbf{0.51} & 0.38 & 0.46 & 0.22 & 0.49 & 0.45 & 0.45 \\
ia-dardic & 491 & 0.23 & 0.18 & 0.19 & 0.18 & \textbf{0.49} & 0.18 & 0.30 & 0.22 & 0.23 & 0.13 \\
ia-east & 1314 & 0.40 & 0.44 & 0.42 & 0.42 & 0.35 & \textbf{0.46} & 0.18 & 0.41 & 0.42 & 0.42 \\
ia-north & 451 & 0.02 & -0.12 & -0.08 & -0.04 & 0.11 & 0.05 & \textbf{0.59} & 0.42 & 0.08 & -0.24 \\
ia-north-west & 1310 & 0.46 & 0.48 & 0.49 & 0.49 & 0.34 & 0.47 & 0.26 & \textbf{0.55} & 0.49 & 0.44 \\
ia-south & 1195 & 0.56 & 0.60 & 0.58 & 0.57 & 0.50 & 0.58 & 0.29 & 0.60 & \textbf{0.62} & 0.58 \\
ia-west & 780 & 0.33 & 0.33 & 0.33 & 0.33 & 0.30 & 0.32 & 0.15 & 0.33 & 0.33 & \textbf{0.36} \\
comtail & 9887 & 0.43 & 0.46 & 0.45 & 0.45 & 0.39 & 0.45 & 0.24 & \textbf{0.48} & 0.45 & 0.42 \\
\bottomrule
\end{tabular*}
\end{table}


\begin{table}[h]
\caption{Language Hindi Urdu QE Ablation Models. Kendall’s Tau ($\tau$) correlations obtained for each language-pair's test set compared with entire COMTAIL test set. The highest correlation value for each test set row is highlighted in bold.}
\label{tbl:abl-language-hin-urd-qe-result}
\small
\centering
\begin{tabular*}{\textwidth} {@{\extracolsep{\fill}}cccccc}
\toprule
test-set & count & comet-kiwi-wmt22-da & ft-qe-enghin & ft-qe-engurd & ft-qe-hinurd \\
\midrule
eng-hin & 571 & 0.57 & \textbf{0.62} & 0.54 & 0.56 \\
eng-urd & 525 & 0.39 & \textbf{0.40} & \textbf{0.40} & \textbf{0.40} \\
hin-urd & 610 & 0.43 & 0.44 & 0.43 & \textbf{0.50} \\
comtail & 9887 & 0.43 & \textbf{0.44} & \textbf{0.44} & \textbf{0.44} \\
\bottomrule
\end{tabular*}

\end{table}




\end{appendices}

\clearpage

\bibliography{sn-bibliography}

\end{document}